\documentclass[]{fairmeta}

\usepackage{wrapfig}
\usepackage{tabularx}
\usepackage{textcomp}
\usepackage{stfloats}
\usepackage{url}
\usepackage{verbatim}
\usepackage{titlesec}
\usepackage{tocloft}
\usepackage{adjustbox}
\usepackage{multirow}
\usepackage{listings}
\usepackage{pifont}
\usepackage{tikz}
\usepackage{comment}
\usepackage{amsmath,amssymb} 
\usepackage{colortbl}  
\usepackage{color}
\usepackage{booktabs} 
\usepackage{hyperref}
\usepackage{graphicx}    
\usepackage{subcaption} 
\usepackage{multirow} 
\usepackage{booktabs} 
\usepackage{subcaption} 
\usepackage{enumitem}
\RequirePackage{xspace}
\makeatletter
\DeclareRobustCommand\onedot{\futurelet\@let@token\@onedot}
\def\@onedot{\ifx\@let@token.\else.\null\fi\xspace}

\makeatother

\definecolor{adptorange}{RGB}{248, 205, 172}
\definecolor{cmpblue}{RGB}{189, 215, 238}
\definecolor{cmpblue}{RGB}{189, 215, 238}

\definecolor{our_red}{RGB}{232,157,160}
\definecolor{our_blue}{RGB}{136,206,230}
\definecolor{our_orange}{RGB}{246,200,168}
\definecolor{our_green}{RGB}{178,211,164}

\definecolor{attn_code0}{RGB}{247,215,200}
\definecolor{attn_code1}{RGB}{238,169,139}
\definecolor{mlp_code0}{RGB}{204,201,221}
\definecolor{mlp_code1}{RGB}{102,95,153}

\definecolor{token_blue}{RGB}{84, 120, 140}

\usepackage{pifont}       
\usepackage{bbding}       
\usepackage{fontawesome}
\usepackage{xspace}

\usepackage{float}

\newlength\savewidth

\newcolumntype{x}[1]{>{\centering\arraybackslash}p{#1pt}}
\newcolumntype{y}[1]{>{\raggedright\arraybackslash}p{#1pt}}
\newcolumntype{z}[1]{>{\raggedleft\arraybackslash}p{#1pt}}

\renewcommand{\paragraph}[1]{\vspace{1mm}\noindent\textbf{#1}}
\usepackage{colortbl}
\usepackage{xcolor}
\usepackage{wrapfig}

\renewcommand{\paragraph}[1]{\vspace{1.25mm}\noindent\textbf{#1}}

\usepackage{algorithm}
\usepackage{listings}

\definecolor{codeblue}{rgb}{0.25, 0.5, 0.5}
\definecolor{codekw}{rgb}{0.35, 0.35, 0.75}
\lstdefinestyle{Pytorch}{
    language = Python,
    backgroundcolor = \color{white},
    basicstyle = \fontsize{9pt}{8pt}\selectfont\ttfamily\bfseries,
    columns = fullflexible,
    aboveskip=1pt,
    belowskip=1pt,
    breaklines = true,
    captionpos = b,
    commentstyle = \color{codeblue},
    keywordstyle = \color{codekw},
}

\definecolor{green}{HTML}{009000}
\definecolor{red}{HTML}{ea4335}

\title{Decision-Native Multimodal Clinical Modeling via Anatomical Grounding and Confidence Calibration}

\renewcommand{\authorlist}{%
  \authorformat[*1, 2]{Juntao Jiang}, %
  \authorformat[*2, 3, 1]{Wenqi Zhong}, %
  \authorformat[*2, 3, 1]{Guangyu Guo}, %
  \authorformat[2]{Dexia Chen}, %
  \authorformat[2]{Yilong Hu}, %
  \authorformat[2, 4]{Yunhao Bai}, \\
  \authorformat[2, 5]{Zhenyu Yi}, %
  \authorformat[2, 3, 1]{Jianwei Xu}, %
  \authorformat[2]{Tony C. W. MOK}, %
  \authorformat[1]{Muxuan Gao}, %
  \authorformat[2]{Ling Zhang}, %
  \authorformat[1]{Yong Liu}, \\
  \authorformat[\dagger1]{Jiangning Zhang}, %
  \authorformat[\dagger2, 3, 1]{Zhongwei Qiu}%
}

\affiliation[1]{Zhejiang University}
\affiliation[2]{DAMO Academy, Alibaba Group\\}
\affiliation[3]{Hupan Lab}
\affiliation[4]{East China Normal University\\}
\affiliation[5]{Shanghai Jiao Tong University}
\contribution[*]{Equal contribution}
\contribution[\dagger]{Corresponding author}

\abstract{
Structured clinical decisions from volumetric computed tomography (CT) require anatomically grounded predictions and calibrated confidence estimates. However, existing 3D medical multimodal models often suffer from uncalibrated autoregressive decoding and shortcut reliance on irrelevant anatomical regions. We introduce \textbf{Radev}, a decision-native multimodal model that integrates anatomy-guided visual compression, direct candidate prediction, and counterfactual confidence supervision. To bridge global context with fine-grained slice details, a frozen 3D encoder extracts reusable organ-level features, which are dynamically selected by the query to attend to 2D slice features within predicted anatomical bounds. The resulting compact visual tokens enable the language backbone to estimate candidate probabilities in a single forward pass without autoregressive decoding. We further propose Counterfactual-Calibrated Training (CCT) to compare predictions under intact and intervened organ representations. CCT encourages confidence to decrease when removing target-organ evidence degrades decision quality, while penalizing excessive confidence drops under unrelated-organ removal. To facilitate rigorous evaluation, we construct \textbf{CT-Verdict}, a comprehensive benchmark spanning binary finding presence, multiple-choice questions, and ordinal severity assessment across both in-distribution and external cohorts. Extensive experiments demonstrate that Radev significantly outperforms state-of-the-art methods in both decision accuracy and confidence calibration, while generalizing robustly under distribution shifts. Project page at \url{https://juntaojianggavin.github.io/projects/Radev/}.
}

\date{\today}

\begin{document}
\thispagestyle{firstheader}
\maketitle
\pagestyle{empty}

\section{Introduction}
\label{sec:intro}

\begin{figure}[t]    
\centering    
\includegraphics[width=0.75\linewidth]{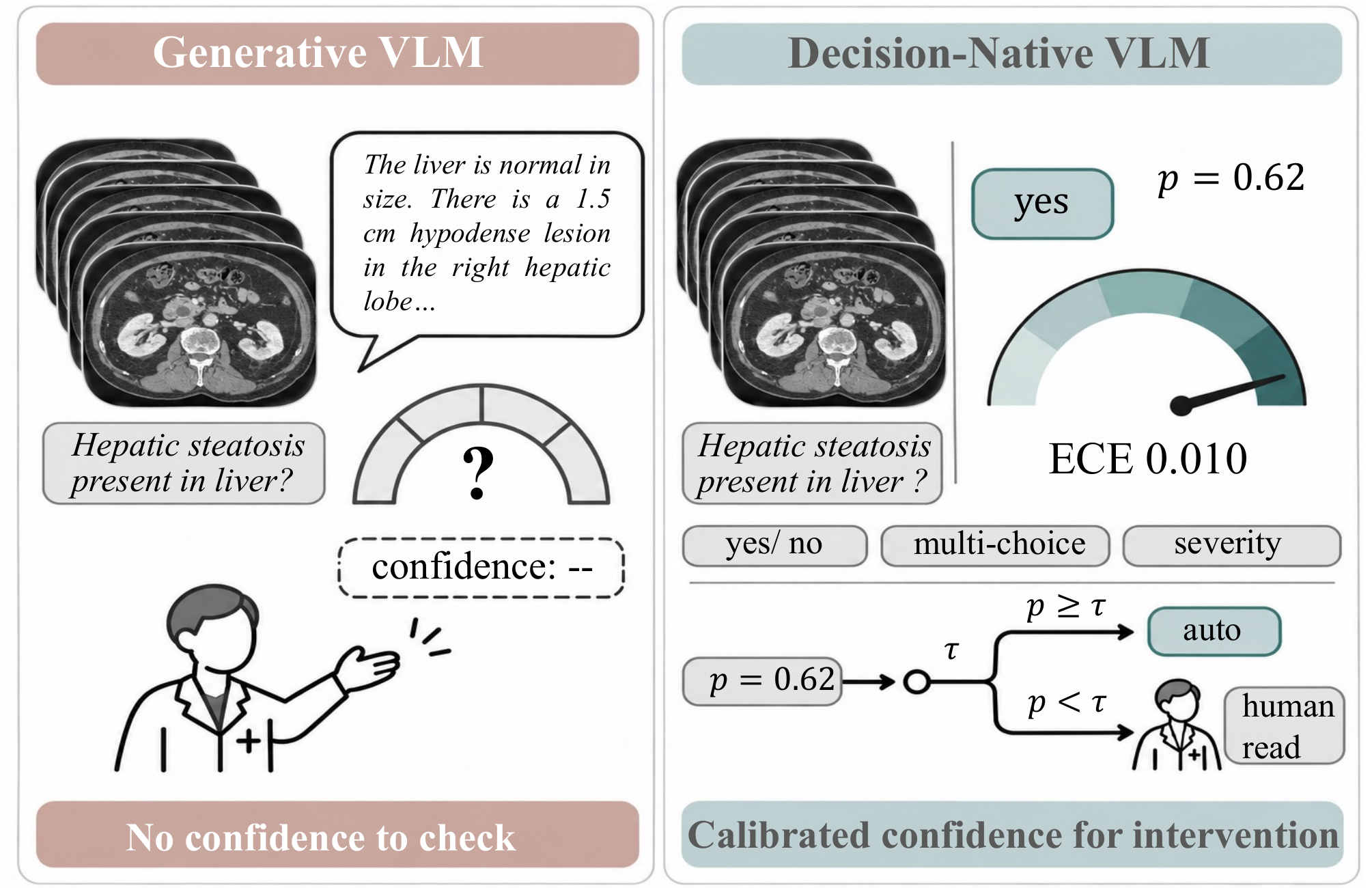} 
\vspace{2mm} 
\caption{\textbf{Generative VLMs vs.\ Decision-Native VLMs.} Conventional generative VLMs are designed for open-ended reasoning and report generation, yet often rely on ungrounded visual context and yield uncalibrated confidence. \textbf{Radev} is a decision-native model tailored for structured, organ-grounded, and confidence-aware clinical judgment.}

\label{fig:teaser}    
\end{figure}

Medical vision language models (VLMs) integrate visual inputs with natural-language queries to support diverse clinical tasks~\citep{jiang2025hulu,qiu2025bridging,xu2026lingshu,yuan2026clinfusion,li2026tumorchain}. In volumetric computed tomography (CT), clinical queries routinely target specific organs or findings to assess their presence, anatomical location, attributes, or severity. Because these judgments carry high clinical stakes, where missed findings delay intervention and false positives trigger unnecessary procedures, clinicians require reliable uncertainty estimates to flag ambiguous cases for human review~\citep{geifman2017selective,angelopoulos2023conformal}. However, overconfident errors undermine this safety net by making erroneous predictions appear trustworthy. Achieving safe clinical deployment therefore demands not only high diagnostic accuracy, but also rigorous {confidence calibration}, ensuring that a model's reported confidence faithfully reflects its empirical accuracy.

Existing volumetric CT VLMs predominantly follow a generative paradigm as shown in Fig.~\ref{fig:teaser}, employing 3D or hybrid encoders to autoregressively generate free-text answers or radiology reports~\citep{wu2025towards,xin2025med3dvlm,li2026mrl,yuan2026clinfusion}. While these models can be adapted to closed-set clinical decisions via text generation or candidate likelihood scoring, this formulation faces three fundamental limitations. First, autoregressive decoding incurs unnecessary computational overhead for structured decisions and often yields poorly calibrated token-level probabilities. Second, detecting localized findings requires fine-grained 2D slice details situated within 3D organ context, where the relevant anatomical evidence shifts dynamically with the query. Third, standard answer-level supervision rewards agreement with ground-truth labels without verifying whether the prediction, or its confidence, is genuinely grounded in the target anatomy rather than spurious background correlations. Building on recent advances in decision modeling~\citep{typesafe_jev_1_14,tang2026omnimed,tommasino2026pathjevdecisionnativevisionlanguagemodelling}, we reformulate structured CT interpretation as a decision process as in Fig.~\ref{fig:teaser} centered on direct candidate probability estimation, query-driven anatomical evidence selection, and counterfactual confidence supervision.

To this end, we introduce \textbf{Radev}, a decision-native multimodal model that unifies anatomically grounded visual compression with counterfactual confidence calibration. Specifically, Radev utilizes a frozen 3D encoder to extract reusable organ-level features, which are projected into a persistent \emph{organ ledger}. Given a clinical query, a text-guided routing mechanism selects the target organs, whose tokens attend to high-resolution 2D slice features within predicted 3D anatomical bounds. The resulting compact visual token sequence preserves global organ context while injecting localized slice evidence specifically for the queried anatomy. Paired with the question and candidate options, these tokens allow the language backbone to estimate candidate probabilities in a single forward pass without autoregressive decoding. Following multimodal alignment and verdict fine-tuning across binary, multiple-choice, and ordinal tasks, we introduce \emph{Counterfactual-Calibrated Training} (CCT). By contrasting model predictions under intact, target-ablated, and control-ablated organ representations, CCT explicitly forces confidence to drop when removing target-organ evidence impairs decision quality, while penalizing unwarranted sensitivity to the removal of unrelated organs.

To enable rigorous evaluation, we construct \textbf{CT-Verdict}, a comprehensive benchmark for structured CT interpretation spanning binary finding presence, multiple-choice location and attribute queries, and ordinal severity assessment. Featuring both in-distribution (IID) and external test cohorts, CT-Verdict establishes a standardized testbed for evaluating accuracy and calibration across task types and distribution shifts. Extensive experiments show that Radev-2B achieves a state-of-the-art overall accuracy of $83.75\%$ on CT-Verdict-IID while substantially improving confidence calibration and out-of-distribution generalization. Our main contributions are summarized as follows:
\begin{itemize}
    \item \textbf{Decision-Native Architecture:} We propose \textbf{Radev}, a structured CT interpretation model that bridges reusable 3D organ ledgers with query-selected 2D slice evidence, enabling single-pass candidate probability estimation without autoregressive decoding.
    \item \textbf{Counterfactual Confidence Supervision:} We design \emph{Counterfactual-Calibrated Training}, an intervention-based training objective that aligns predictive confidence with causal anatomical evidence by contrasting intact, target-ablated, and control-ablated organ states.
    \item \textbf{Benchmark and Empirical Validation:} We release \textbf{CT-Verdict}, a multi-task volumetric CT benchmark with internal and external cohorts, and demonstrate that Radev consistently outperforms existing generative VLMs in diagnostic accuracy, confidence calibration, and robustness to distribution shifts.
\end{itemize}

\section{Related Work}
\label{sec:related_work}

\subsection{Decision Model}
Direct prediction over candidate answers has been explored in language and vision language modeling. BERT~\citep{devlin2019bert} scores candidate continuations for multiple-choice inference, while LXMERT~\citep{tan2019lxmert} predicts answers from a predefined vocabulary conditioned on an image and a question. These discriminative formulations provide an established basis for making decisions without free-text generation. Recent decision-native language models revisit this principle through interfaces that accept a task description and an explicit candidate set and return candidate probabilities. Jev~\citep{typesafe_jev_1_14} is one such system, mapping a typed state and discrete candidates to a decision distribution in a single forward pass. Open replicas fine-tune general-purpose backbones with decision heads or pointer readouts like Kev~\citep{kev2026}, decider~\citep{marosi2026decider}, OneJev~\citep{onejev2026}, ReJev~\citep{rejev2026}, which retain the autoregressive backbone but score candidates. 

The decision interface has recently been extended to vision and other modalities. Intern-Decision~\citep{zhu2026intern} couples Qwen3.5 backbones to return an answer distribution over text states with optional images in one forward pass. OmniJev~\citep{omnijev2026} and Clef~\citep{clefomni2026} accept text, image, audio, and video. OmniMed-Jev~\citep{tang2026omnimed} couples a vision--language backbone with decision-native calibration to answer questions
across different medical modalities.
PathJev~\citep{tommasino2026pathjevdecisionnativevisionlanguagemodelling} trains a decision head on histopathology patches and
shows that candidate scoring with a dedicated head outperforms reading
probabilities off a generative model's language head. However, all of them ingest 2D
images and none consumes 3D volumetric studies, which are ubiquitous and indispensable in clinical medical imaging.
\subsection{Medical Vision Language Model}
VLMs are now widely used in medical imaging, answering report generation and visual question-answering. The domain-specific nature of medical images has given medical VLMs a distinct set of design choices. One family retains standard 2D architectures. LLaVA-Med~\citep{li2023llava} establishes this paradigm via concept alignment and instruction tuning. HealthGPT~\citep{lin2025healthgpt} introduces H-LoRA plugins to unify medical comprehension and generation. Despite these advances, their visual inductive bias remains 2D without volumetric supervision. Consequently, volumetric images like CT studies are reduced to 2D slices, a compromise formalized by Hulu-Med~\citep{jiang2025hulu}, which processes slices as patch sequence via rotary-position-adaptive encoding. To reduce inter-slice redundancy, CARVE~\citep{yi2026carve} frames training-free token compression as a budget-constrained 2.5D allocation problem to retain salient visual evidence.

A second family integrates 3D encoders, addressing two key bottlenecks: volumetric token explosion and the lack of web-scale 3D corpora. To tame token complexity, RadFM~\citep{wu2025towards} relies on interleaved 2D/3D generative pretraining, M3D-LaMed~\citep{bai2024m3d} enforces a fixed token budget via a spatial-pooling perceiver, and Med3DVLM~\citep{xin2025med3dvlm} substitutes dense 3D attention with efficient decomposed convolutions. To compensate for data scarcity via rich supervision, CT-CHAT~\citep{hamamci2026generalist} reuses a frozen CT-CLIP encoder, while Merlin~\citep{blankemeier2024merlin} pretrains a 3D ResNet jointly on reports and EHR codes. Bridging both worlds, hybrid models retain 2D priors while injecting 3D cues modularly: OmniCT~\citep{lin2026omnict} and TIF-GRPO~\citep{lin2026regulating} combine 3D positional encodings with mask-guided organ token aggregation, and ClinFusion~\citep{yuan2026clinfusion} fuses 3D volumes with 2D foundation features via local cross-attention. Despite these advances, decision VLMs tailored for 3D medical imaging remain largely unexplored.

\section{Method}
\label{sec:method}

\subsection{Task Definition and Overview}
\label{sec:overview}
\begin{figure*}[t!]
    \centering
    \includegraphics[width=\linewidth]{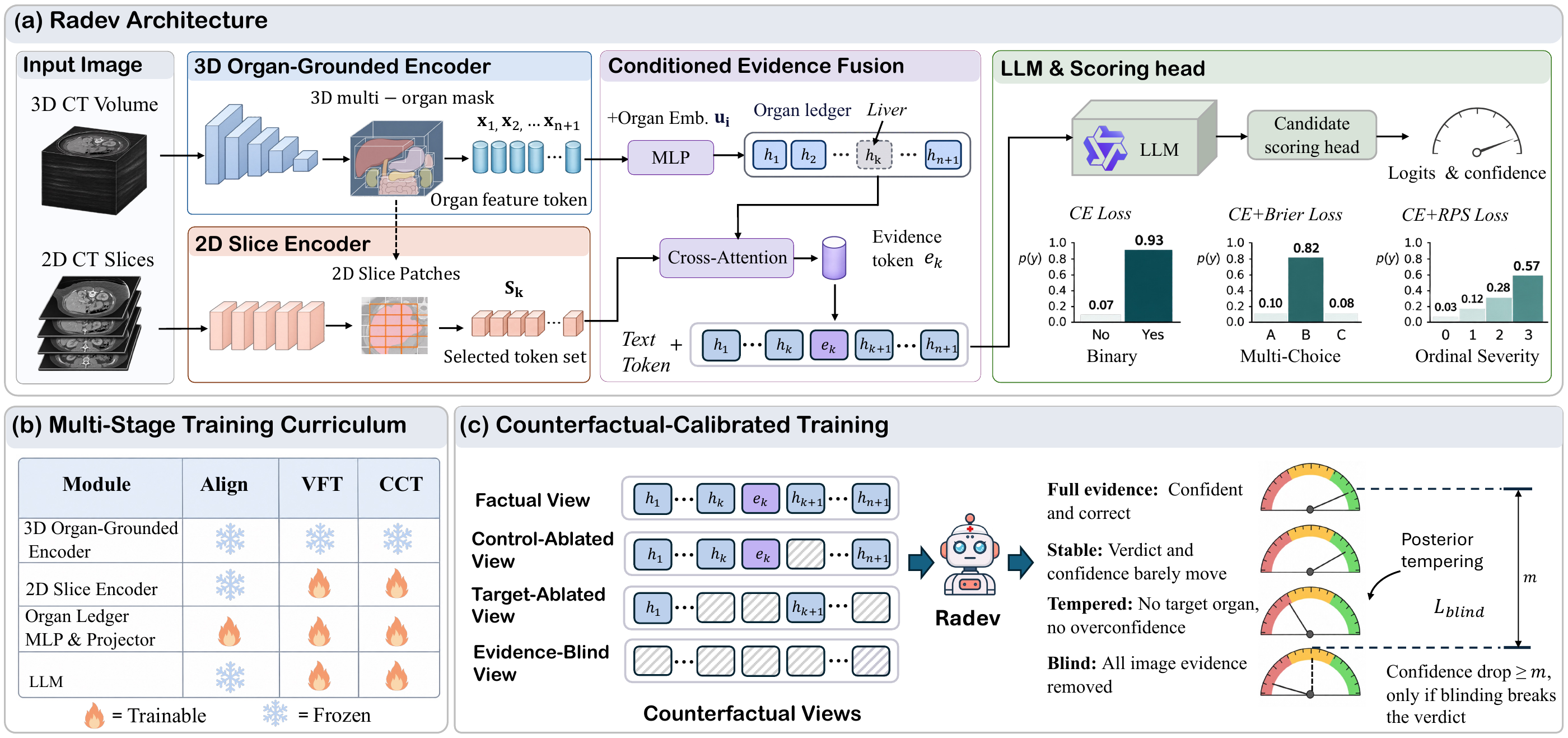}
    \vspace{1mm} 
    \caption{\textbf{Framework of the proposed Radev.} (a) The Organ-Grounded Visual Encoder uses organ masks and learned mask queries to extract 3D organ tokens. Question-selected organ tokens attend to 2D slice features within their anatomical bounds. The language backbone and candidate scoring head predict an answer distribution in a single forward pass. (b) Training proceeds through projector–evidence alignment, verdict fine-tuning, and Counterfactual-Calibrated Training (CCT). (c) CCT constructs factual, target-ablated, control-ablated, and evidence-blind views by intervening on visual features. The first three views receive task supervision, while the evidence-blind view receives a confidence-margin penalty.}
    \label{fig:architecture}
\end{figure*}
Given a 3D CT volume $V$ and a clinical question $q$ with candidate answers $\mathcal{C}(q)$, Radev predicts a distribution $p_\theta(c\mid V,q)$ over the candidates. We consider binary finding presence, multiple-choice questions about finding location or attributes, and ordinal severity assessment. Each question has exactly one reference answer.

Radev combines an Organ-Grounded Visual Encoder with a Qwen3.5 language backbone and a candidate scoring head (Fig.~\ref{fig:architecture}). The encoder uses organ masks to guide 3D organ-token readout and region-restricted attention to incorporate question-relevant 2D slice evidence. Projector-evidence alignment and verdict fine-tuning (VFT) directly supervise candidate probabilities. Counterfactual-Calibrated Training (CCT) then combines task supervision across intact and ablated representations with a confidence constraint on an evidence-blind view.

\subsection{Organ-Grounded Visual Encoder}
\label{sec:encoding}

\noindent\textbf{Mask-guided organ tokenization.}
Given a 3D CT volume $V$, the frozen 3D branch provides multi-scale visual feature tokens and organ masks $\{\Omega_i\}_{i=1}^n$ for $n$ organ classes. Each mask $\Omega_i$ restricts spatial support to $\mathbf{T}_i$, the multi-scale 3D feature tokens whose spatial cells overlap $\Omega_i$. An organ-associated mask token $\mathbf{m}_i$, serving as a learned readout query, aggregates $\mathbf{T}_i$ via shared multi-head attention, an organ-specific linear projection $\mathbf{W}_i$, and $\ell_2$ normalization to yield the $d$-dimensional organ feature $\mathbf{x}_i$,
\begin{equation}
\mathbf{f}_i=\mathbf{W}_i\,\mathrm{MHA}(\mathbf{m}_i,\mathbf{T}_i,\mathbf{T}_i),
\qquad
\mathbf{x}_i=\frac{\mathbf{f}_i}{\|\mathbf{f}_i\|_2}.
\label{eq:readout}
\end{equation}
A separate global readout extracts a whole-volume feature $\mathbf{x}_{n+1}\in\mathbb{R}^d$. An organ that is undetected, truncated by the field of view, or not fully contained in any processing window uses a shared learned feature placeholder in place of $\mathbf{x}_i$ before projection. A shared MLP maps all $n+1$ features into the language embedding space, and learned slot embeddings $\mathbf{u}_i$ identify each anatomical or global slot,
\begin{equation}
\mathbf{h}_i=\mathrm{MLP}(\mathbf{x}_i)+\mathbf{u}_i,
\qquad i=1,\dots,n+1.
\label{eq:ledger}
\end{equation}
The resulting \emph{organ ledger} retains all $n+1$ slots in a fixed order. Constructed from $V$ alone, it is reusable across questions, with backbone and windowing details provided in Sec.~\ref{sec:impl}.

\noindent\textbf{Question-conditioned evidence fusion.}
The native Qwen3.5 vision encoder extracts tokens $\mathbf{S}=(\mathbf{s}_1,\dots,\mathbf{s}_L)$ from uniformly sampled axial slices. A deterministic text-based mapping selects the organ set $\mathcal{B}(q)$ relevant to $q$, using $\mathcal{B}(q)=\{n+1\}$ when no organ is bound. For each selected organ $k$, $\mathbf{S}_k$ contains slice tokens whose spatial centers lie within the 3D bounding box $R_k$ of $\Omega_k$, while the global fallback uses $\mathbf{S}_{n+1}=\mathbf{S}$. Using the projected organ token $\mathbf{h}_k$ as the query and $\mathbf{S}_k$ as keys and values, residual attention combines organ context with regional slice detail into an evidence token,
\begin{equation}
\mathbf{e}_k=\mathbf{h}_k+
\mathrm{MHA}(\mathbf{h}_k,\mathbf{S}_k,\mathbf{S}_k),
\qquad k\in\mathcal{B}(q).
\label{eq:evidence}
\end{equation}
The question thus selects organs, mask bounds restrict eligible slice tokens, and learned attention aggregates their content. The attention output projection is zero-initialized, and empty token sets $\mathbf{S}_k$ return $\mathbf{e}_k=\mathbf{h}_k$ without a 2D update. Each evidence token $\mathbf{e}_k$ is inserted immediately after $\mathbf{h}_k$, preserving all base ledger slots. The language backbone receives $n+1+K$ visual tokens, where $K=|\mathcal{B}(q)|$, instead of raw slice tokens.

\subsection{Decision-Native Prediction and Supervision}
\label{sec:head}
\label{sec:training}
\label{sec:twostage}

Qwen3.5 processes the interleaved visual tokens together with the question and candidate answers. Each candidate $c$ is represented by a single vocabulary token. Its logit $z_c$ at the final prompt position is normalized over the candidate set,
\begin{equation}
 p_\theta(c\mid V,q)=
 \frac{\exp z_c}{\sum_{c'\in\mathcal{C}(q)}\exp z_{c'}}.
\label{eq:head}
\end{equation}
The predicted answer and confidence are
\begin{equation}
\hat c=\arg\max_{c\in\mathcal{C}(q)}p_\theta(c\mid V,q),
\qquad \hat p=p_\theta(\hat c\mid V,q).
\label{eq:task_prediction}
\end{equation}
This readout requires one language-backbone forward pass per question without autoregressive answer decoding. Training begins with projector--evidence alignment, followed by verdict fine-tuning (VFT). Both stages directly supervise the candidate distribution using the same prompt and readout.

For a batch of $N$ examples, $\mathcal{L}_{\mathrm{task}}=N^{-1}\sum_{i=1}^{N}\ell_i$. Let $p_{i,k}$ denote the probability of candidate $k$, $y_i$ the reference index, and $K_i$ the number of candidates. The task-specific losses are
\begin{equation}
\begin{aligned}
\ell_i^{\mathrm{bin}}&=-\log p_{i,y_i},\\
\ell_i^{\mathrm{ord}}&=-\log p_{i,y_i}
 +\lambda_{\mathrm{rps}}\sum_{j=1}^{K_i-1}
 (\widehat F_{i,j}-F_{i,j})^2,\\
\ell_i^{\mathrm{mc}}&=-\log p_{i,y_i}
 +\lambda_{\mathrm b}\sum_{k=1}^{K_i}
 (p_{i,k}-\mathbb I(k=y_i))^2.
\end{aligned}
\label{eq:loss-kinds}
\end{equation}
Binary questions use cross-entropy. Ordinal questions additionally use the ranked probability score, with $\widehat F_{i,j}=\sum_{k=1}^{j}p_{i,k}$ and $F_{i,j}=\mathbb I(y_i\le j)$, to account for the ordering of severity levels. Multiple-choice questions add the Brier score, which measures squared error against the one-hot reference distribution.

\subsection{Counterfactual-Calibrated Training}
\label{sec:cct}

Supervision on intact inputs does not directly constrain confidence when anatomical evidence is withheld. Initialized from the VFT checkpoint, Counterfactual-Calibrated Training (CCT) combines supervision across different evidence conditions with an explicit confidence constraint on an evidence-blind view.

\noindent\textbf{Counterfactual views.}
CCT replaces the 3D features of selected slots with the shared missing-feature placeholder before the MLP in Eq.~\eqref{eq:ledger} and disables their 2D evidence updates. Each affected evidence token then reduces to its intervened ledger token, $\mathbf{e}_k=\mathbf{h}_k$, preserving the interleaved token layout. For a volume $V$ and question $q$ mapped to organ set $\mathcal{B}(q)$, CCT evaluates four views sharing the same question text and gold answer. The \emph{factual} view retains all extracted 3D features and 2D evidence updates. The \emph{target-ablated} view replaces all bound organs in $\mathcal{B}(q)$ together with the global slot $n{+}1$, while retaining unbound organs. The \emph{control-ablated} view replaces one randomly sampled present organ outside $\mathcal{B}(q) \cup \{n{+}1\}$ and never ablates the global slot. The \emph{evidence-blind} view replaces all $n{+}1$ visual slots with placeholders and disables all 2D updates, removing every image-derived feature while preserving the placeholder token sequence. Infeasible target or control views fall back to factual readouts in the task average, whereas unbound questions still admit the evidence-blind view.

\noindent\textbf{Multi-view confidence supervision.}
During CCT, $\mathcal{L}_{\mathrm{task}}$ averages the losses in Eq.~\eqref{eq:loss-kinds} across the factual, target-ablated, and control-ablated readouts using the same reference answer. This applies the same scoring rules under different evidence conditions without prescribing synthetic soft targets. The evidence-blind view is excluded from task supervision and instead receives a confidence-margin penalty. Let $c$ and $\breve c$ denote the maximum candidate probabilities under the factual and evidence-blind views, respectively. Their decision scores $r$ and $\breve r$ are exact-match correctness for binary and multiple-choice questions, and one minus the absolute level error divided by $K_i-1$ for ordinal questions with $K_i$ levels. For a single example, the auxiliary loss is
\begin{equation}
  \mathcal{L}_{\mathrm{blind}}
  = \mathbb{I}(r>\breve{r})\bigl[m_{\mathrm{blind}}-(\operatorname{sg}[c]-\breve{c})\bigr]_+,
  \label{eq:cct-blind}
\end{equation}
where $[x]_+=\max(0,x)$. The indicator activates the loss only when withholding all image-derived features reduces the decision score. The hinge encourages blind confidence to be at least $m_{\mathrm{blind}}$ below the factual confidence. The stop-gradient $\operatorname{sg}[\cdot]$ blocks the direct blind-loss gradient through $c$, while gradients through the blind prediction remain active. The overall CCT objective is
\begin{equation}
  \mathcal{L}_{\mathrm{CCT}}
  = \mathcal{L}_{\mathrm{task}}
   + \lambda_{\mathrm{blind}}\mathcal{L}_{\mathrm{blind}},
  \label{eq:cct-total}
\end{equation}
with $\lambda_{\mathrm{blind}}=0.2$ and $m_{\mathrm{blind}}=0.20$.
\section{CT-Verdict Benchmark}
\label{sec:benchmark}

CT-Verdict pairs 3D CT volumes with structured clinical questions and candidate answers to evaluate decision performance on in-distribution and external data. It covers the three task types defined in Sec.~\ref{sec:overview}.

\subsection{Question Construction and Verification}

Radiology reports are first parsed into structured finding records containing the finding, its associated organ or anatomical region, and its polarity, with optional attributes such as severity, size, number, laterality, and relations to neighboring structures. Fixed synthesis rules convert these records into question text, candidate sets, and reference labels. Binary questions use No/Yes answers, multiple-choice questions contain two to five candidates, and ordinal questions use five severity levels from 0 to 4.

Two independent LLM readers answer each question using its source sentence, and their answers are compared with the rule-derived label. Disagreements are reviewed by a third reader. A random sample is additionally audited by a human reader. Questions that fail the reader-consistency criterion are discarded rather than relabeled. Question synthesis and label verification are separate stages, with the former determining the question text and reference labels and the latter checking their agreement with the source reports. 

\subsection{Data Sources and Splits}

\textbf{CT-Verdict-IID.} Merlin-CT~\citep{blankemeier2024merlin} supplies 25{,}270 contrast-enhanced and non-enhanced abdominal CT studies with paired reports for training and in-distribution evaluation. All questions from the same study are assigned to the same split. The test set contains 56{,}825 questions from 5{,}080 studies, comprising 37{,}925 binary (\textsc{noul}), 16{,}210 multiple-choice (\textsc{choice}), and 2{,}690 ordinal (\textsc{score}) questions across 17 organ systems and 30 clinical findings.

\noindent\textbf{CT-Verdict-OOD.} The external test set uses 421 multi-centre CT studies with reports from AMOS-MM~\citep{ji_2024_10992155}. It contains 2{,}055 questions, comprising 1{,}195 binary, 801 multiple-choice, and 59 ordinal questions across 12 organ systems and 24 clinical findings.

\subsection{Evaluation Protocol}

For binary and multiple-choice questions, we report per-question accuracy (Acc), balanced accuracy (mBAcc), and area under the ROC curve (mAUC). The latter two metrics are macro-averaged across the disease categories represented in the questions. For ordinal severity, we report mean absolute error (MAE), the normalized ranked probability score (KRPS)~\citep{epstein1969scoring}, and quadratic weighted kappa (QWK)~\citep{cohen1968weighted}, measuring prediction error, ordinal probabilistic accuracy, and chance-corrected ordinal agreement, respectively. We also evaluate probability calibration using top-label expected
calibration error (ECE)~\citep{guo2017calibration} and the multiclass Brier score~\citep{glenn1950verification}. To disentangle miscalibration from discrimination, we further
apply the Murphy decomposition and report
reliability (REL) and resolution (RES)~\citep{murphy1973new}.

\section{Experiment}

           \begin{table*}[h]
            \centering
            \caption{Comparison of representative general and medical VLMs on the \textit{CT-Verdict-IID}. Accuracy (Acc.) is per-question exact-match accuracy. mBAcc and mAUC are macro means over the finding diseases. Score MAE, KRPS, and QWK are computed over levels 0--4. \textbf{Bold} marks the best result in each 
        column and \underline{underline} the runner-up.}
            \label{tab:model_comparison}
            \begingroup
         \small
            \setlength{\tabcolsep}{2.8pt}
            \renewcommand{\arraystretch}{0.83}
            \resizebox{\textwidth}{!}{%
            \begin{tabular}{@{}lccccccccccccccc@{}}
                \toprule
                \multirow{2}{*}{\textbf{Model}}
                & \multicolumn{5}{c}{\textbf{Noul}}
                & \multicolumn{5}{c}{\textbf{Choice}}
                & \multicolumn{5}{c}{\textbf{Score}}\\                                                                                                                                                                                                                                                         
                \cmidrule(lr){2-6}\cmidrule(lr){7-11}\cmidrule(lr){12-16}
                & Acc$\uparrow$ & mBAcc$\uparrow$ & mAUC$\uparrow$ & ECE$\downarrow$ & Brier$\downarrow$
                & Acc$\uparrow$ & mBAcc$\uparrow$ & mAUC$\uparrow$ & ECE$\downarrow$ & Brier$\downarrow$
                & MAE$\downarrow$ & KRPS$\downarrow$ & QWK$\uparrow$ & ECE$\downarrow$ & Brier$\downarrow$\\
                \midrule
                \multicolumn{16}{@{}l}{\textit{General VLMs}}\\
                \addlinespace[0.3pt]
                LLaVA-Med-7B~\citep{li2023llava} & 39.94 & 50.61 & 61.31 & 0.35 & 0.71 & 19.69 & 49.82 & 50.15 & 0.42 & 1.05 & 0.75 & 0.14 & 0.15 & 0.19 & 0.93\\
                \addlinespace[0.3pt]
                RadFM~\citep{wu2025towards} & 58.88 & 50.03 & 51.72 & 0.11 & 0.47 & 39.36 & 49.92 & 50.42 & 0.25 & 0.89 & 0.93 & 0.21 & 0.07 & 0.54 & 1.19\\
                \addlinespace[0.3pt]
                CT-Chat~\citep{hamamci2026generalist} & 41.92 & 50.09 & 52.07 & 0.34 & 0.71 & 22.30 & 50.21 & 52.22 & 0.41 & 1.11 & 0.96 & 0.18 & -0.08 & 0.46 & 1.03\\
                \addlinespace[0.3pt]
                M3D-Phi3-4B~\citep{bai2024m3d} & 47.48 & 54.55 & 57.85 & 0.50 & 1.01 & 9.71 & 50.01 & 50.95 & 0.85 & 1.72 & 0.99 & 0.21 & 0.11 & 0.51 & 1.13\\                                                                                                                                                  
                \addlinespace[0.3pt]
                M3D-Llama2-7B~\citep{bai2024m3d} & 52.31 & 50.45 & 52.27 & 0.18 & 0.61 & 8.72 & 50.01 & 50.39 & 0.81 & 1.67 & 1.00 & 0.22 & -0.01 & 0.60 & 1.32\\
                \addlinespace[0.3pt]
                Merlin~\citep{blankemeier2024merlin} & 66.56 & 56.06 & 79.70 & 0.16 & 0.43 & 38.22 & 50.56 & 61.49 & 0.14 & 0.77 & 0.70 & 0.14 & 0.13 & 0.24 & 0.89\\
                \addlinespace[0.3pt]
                HealthGPT-M3~\citep{lin2025healthgpt} & 43.14 & 49.79 & 55.40 & 0.32 & 0.71 & 23.12 & 50.19 & 52.93 & 0.25 & 0.89 & 1.04 & 0.18 & 0.06 & 0.12 & 0.83\\
                \addlinespace[0.3pt]
                Med3DVLM~\citep{xin2025med3dvlm} & 63.13 & 50.44 & 54.94 & 0.28 & 0.65 & 28.74 & 50.89 & 52.75 & 0.52 & 1.21 & 0.80 & 0.17 & 0.29 & 0.44 & 1.04\\
                \addlinespace[0.3pt]
                InternVL3.5-8B~\citep{wang2025internvl3} & 43.94 & 53.67 & 72.44 & 0.24 & 0.57 & 20.23 & 49.34 & 49.62 & 0.48 & 1.16 & 1.05 & 0.17 & -0.02 & 0.09 & 0.78\\
                \addlinespace[0.3pt]
                Hulu-med-4B~\citep{jiang2025hulu} & 50.71 & 57.72 & 74.71 & 0.13 & 0.47 & 30.31 & 52.86 & 64.08 & 0.27 & 0.94 & 1.06 & 0.17 & 0.24 & 0.11 & 0.80\\
                \addlinespace[0.3pt]
                Hulu-med-7B~\citep{jiang2025hulu} & 64.67 & 61.73 & 78.71 & 0.07 & 0.41 & 29.00 & 56.00 & 67.44 & 0.17 & 0.83 & 1.39 & 0.24 & 0.22 & 0.25 & 0.95\\
                \addlinespace[0.3pt]
                Qwen3.5-4B~\citep{qwen3.5} & 53.16 & 58.44 & 75.02 & 0.15 & 0.51 & 29.69 & 52.54 & 64.87 & 0.29 & 0.92 & 1.12 & 0.19 & 0.23 & 0.14 & 0.85\\                                                                                                                                                     
                \addlinespace[0.3pt]
                Qwen3.5-9B~\citep{qwen3.5} & 57.90 & 63.08 & 76.58 & 0.09 & 0.44 & 36.64 & 54.52 & 67.60 & 0.24 & 0.79 & 1.27 & 0.22 & 0.19 & 0.16 & 0.88\\
                \addlinespace[0.3pt]
                OmniCT-3B~\citep{lin2026omnict} & 44.79 & 52.28 & 56.73 & 0.16 & 0.54 & 23.79 & 50.25 & 55.46 & 0.37 & 1.00 & 1.17 & 0.21 & 0.07 & 0.30 & 0.93\\
                \addlinespace[0.3pt]
                  OmniCT-7B~\cite{lin2026omnict} & 44.86 & 51.86 & 53.20 & 0.18 & 0.56 & 19.44 & 50.57 & 54.89 & 0.33 & 0.97 & 0.88 & 0.15 & 0.02 & 0.19 & 0.81\\
                \addlinespace[0.3pt]
                HealthGPT-Pro-4B~\citep{lin2025healthgpt} & 70.96 & 65.90 & 76.68 & 0.10 & 0.41 & 37.50 & 55.73 & 69.57 & 0.07 & 0.74 & 1.05 & 0.17 & 0.30 & 0.13 & 0.80\\                                                                                                                                      
                \addlinespace[0.3pt]
                HealthGPT-Pro-8B~\citep{lin2025healthgpt} & 62.03 & 68.04 & 77.31 & 0.02 & 0.45 & 35.93 & 55.89 & 67.80 & 0.08 & 0.76 & 0.93 & 0.15 & 0.34 & 0.09 & 0.77\\
                \addlinespace[0.3pt]
                ClinFusion-8B~\citep{yuan2026clinfusion} & 65.66 & 56.58 & 64.68 & 0.08 & 0.44 & 37.65 & 51.26 & 56.51 & 0.11 & 0.72 & 1.02 & 0.17 & 0.22 & 0.09 & 0.81\\
                \addlinespace[0.3pt]
                GPT-5.4~\citep{openai_gpt54} & 70.23 & \underline{69.24} & -- & -- & -- & 45.92 & \textbf{64.31} & -- & -- & -- & 0.75 & -- & 0.46 & -- & --\\
                \midrule
               


                \multicolumn{16}{@{}l}{\textit{Decision VLMs}}\\
                \addlinespace[0.3pt]                                                                                                                                                                                                                                                                                 
                OmniMed-Jev~\citep{tang2026omnimed} & 59.68 & 50.55 & 50.66 & 0.39 & 0.78 & 36.06 & 51.39 & 57.91 & 0.37 & 0.93 & 0.79 & 0.17 & 0.11 & 0.31 & 0.93\\
                 \addlinespace[0.3pt]
                  Intern-Decision-2B~\cite{zhu2026intern} & 56.81 & 60.04 & 70.13 & 0.02 & 0.49 & 38.05 & 58.92 & 68.82 & 0.03 & 0.72 & 0.57 & 0.10 & 0.40 & 0.07 & 0.68\\
                \addlinespace[0.3pt]
                 Intern-Decision-4B~\cite{zhu2026intern} & 64.57 & 60.42 & 72.04 & 0.03 & 0.43 & 45.49 & 57.37 & 65.17 & \textbf{0.02} & 0.68 & 0.52 & 0.10 & 0.35 & 0.13 & 0.73\\
                \addlinespace[0.3pt]
                Jev-1.14~\citep{typesafe_jev_1_14} & 71.39 & 65.14 & 79.31 & 0.04 & 0.37 & 50.99 & 58.52 & 71.78 & 0.02 & 0.63 & 0.60 & 0.12 & 0.36 & 0.21 & 0.79\\
                \addlinespace[0.3pt]
                \midrule
                \textbf{Radev-0.8B(Ours)} & \underline{88.20} & 66.59 & \underline{85.48} & \underline{0.01} & \underline{0.17} & \underline{68.61} & 57.73 & \underline{78.19} & 0.02 & \underline{0.44} & \underline{0.40} & \underline{0.06} & \underline{0.53} & \underline{0.03} & \underline{0.42}\\
                \addlinespace[0.3pt]
                \textbf{Radev-2B(Ours)} & \textbf{90.08} & \textbf{71.79} & \textbf{87.79} & \textbf{0.01} & \textbf{0.15} & \textbf{71.33} & \underline{62.06} & \textbf{81.91} & \underline{0.02} & \textbf{0.41} & \textbf{0.39} & \textbf{0.06} & \textbf{0.54} & \textbf{0.02} & \textbf{0.41}\\
                \bottomrule
            \end{tabular}%
            }
            \endgroup
        \end{table*}
                
 \subsection{Implementation Details}
    \label{sec:impl}

\noindent\textbf{Model configuration.}
We instantiate the 3D branch with the frozen RADAR segmentation-and-encoding network~\citep{zhang2026expert}, which is never trained on the source of \textit{CT-Verdict} test set. It provides $256$-dimensional organ readouts ($d=256$) for $36$ organ slots and one global slot ($n=36$). Each organ is read out from the first processing window that contains it in full. The global feature uses its own query, attention module, and projection to aggregate all multi-scale tokens from a separate unwindowed pass over a differently preprocessed copy of the volume. Both the 2D vision encoder and the language backbone are from Qwen3.5~\citep{qwen3.5}.

\noindent\textbf{Supervised configuration.}
During projector--evidence alignment, the Qwen3.5 language backbone and native 2D vision encoder are frozen while the organ projector and evidence attention are trained. VFT jointly updates the language backbone, 2D vision encoder, and bridging modules. The 3D branch remains frozen throughout training. In Eq.~\eqref{eq:loss-kinds}, we use $\lambda_{\mathrm b}=0.1$ and $\lambda_{\mathrm{rps}}=0.5$.

    \noindent\textbf{Training.} All models run on four 96-GB H20 GPUs with
    study-level data-parallel sharding, so all questions of one study are
    handled by one rank. Both supervised phases use an effective batch of 32, AdamW~\citep{ilya2019decoupled}  with a fresh optimizer
    state at each phase. We train for  $1{,}000$ steps at $10^{-3}$ learning rate in alignment phase, an epoch at $2\times10^{-5}$ in VFT phase and 700 steps at $4\times10^{-6}$ in CCT phase.

    \noindent\textbf{Inference.}   Our model is read out
    exactly as it is trained, through the restricted softmax of
    Eq.~\eqref{eq:head}.  For baselines, likelihood readout softmaxes
the summed token log-probabilities of $K$ teacher-forced candidate
strings under each model's chat template to obtain the answer and full candidate distribution. Decision models expose no
generation path, whereas GPT-5.4 returns text without candidate
log-probabilities. Slice counts follow each baseline’s official 3D configuration where available; otherwise, they default to 32, or are capped at the model's maximum input limit if it only supports a few images.

\subsection{Results}
\noindent\textbf{Main Results on CT-Verdict-IID.} 
Table~\ref{tab:model_comparison} compares Radev against representative general and decision-native VLMs across two readout modes. First, Radev establishes a decisive lead
across all three tasks: even our 2B variant achieves 83.75\% overall accuracy,
outperforming the strongest open-source baselines, as 2D decision-native
models lack volumetric context and generative VLMs struggle with ordinal severity.
Closed-source models generalize better than open-source baselines yet still trail Radev
by $\sim$19\%, underscoring the necessity of domain-specific grounding.  Beyond raw accuracy, a more critical requirement for clinical deployment is whether a model's reported confidence is trustworthy. Even the most accurate open- and closed-source baselines remain severely overconfident and miscalibrated, confirming that task accuracy alone does not guarantee reliable uncertainty estimates. In contrast, Radev achieves the lowest Brier scores across all task families and drives top-label ECE down to the $0.01\text{--}0.02$ range. This ensures that Radev's predicted probabilities can be directly thresholded for selective prediction and clinical abstention.


     \begin{table}[h]
                    \centering
                    \caption{Comparison of representative general and medical VLMs on the \textit{CT-Verdict-OOD}.}
                    \label{tab:model_comparison_2} 
                    \begingroup
                 \small
                    \setlength{\tabcolsep}{12pt}
                    \renewcommand{\arraystretch}{0.8}                                                                                                                                                                                                                                                         
                    \begin{tabular}{@{}lcccccc@{}}
                        \toprule
                        \multirow{2}{*}{\textbf{Model}}
                        & \multicolumn{2}{c}{\textbf{Noul}}
                        & \multicolumn{2}{c}{\textbf{Choice}}
                        & \multicolumn{2}{c}{\textbf{Score}} \\
                        \cmidrule(lr){2-3}
                        \cmidrule(lr){4-5}                                                                                                                                                                                                                                                                     
                        \cmidrule(lr){6-7}
                        & Acc$\uparrow$ & mAUC$\uparrow$
                        & Acc$\uparrow$ & mAUC$\uparrow$
                        & MAE$\downarrow$ & KRPS$\downarrow$ \\
                        \midrule
     
                        \multicolumn{7}{@{}l}{\textit{General VLMs}} \\
                        \addlinespace[0.3pt]
                        LLaVA-Med-7B~\citep{li2023llava} & 35.80 & 55.80 & 17.60 & 58.50 & 0.653 & 0.152 \\
                        \addlinespace[0.3pt]
                        RadFM~\citep{wu2025towards} & 55.70 & 43.50 & 55.40 & 46.00 & 0.582 & 0.130 \\
                        \addlinespace[0.3pt]
                        CT-Chat~\citep{hamamci2026generalist} & 40.00 & 39.70 & 8.90 & 51.30 & 1.440 & 0.229 \\
                        \addlinespace[0.3pt]
                        M3D-Llama2-7B~\citep{bai2024m3d} & 41.80 & 53.10 & 15.70 & 51.40 & 0.993 & 0.222 \\
                        \addlinespace[0.3pt]
                        Merlin~\citep{blankemeier2024merlin} & 61.30 & 69.90 & 11.20 & 62.70 & 0.501 & 0.141 \\
                        \addlinespace[0.3pt]
                        HealthGPT-M3~\citep{lin2025healthgpt} & 42.30 & 47.90 & 22.60 & 50.60 & 1.125 & 0.185 \\
                        \addlinespace[0.3pt]
                        Med3DVLM~\citep{xin2025med3dvlm} & 56.40 & 60.30 & 9.10 & 53.90 & 0.910 & 0.197 \\
                        \addlinespace[0.3pt]
                        InternVL3.5-8B~\citep{wang2025internvl3} & 42.00 & 64.70 & 18.10 & 56.70 & 0.945 & 0.159 \\
                        \addlinespace[0.3pt]
                        Hulu-med-7B~\citep{jiang2025hulu} & 64.90 & 66.20 & 19.60 & 71.40 & 1.417 & 0.237 \\
                        \addlinespace[0.3pt]
                       Qwen3.5-9B~\citep{qwen3.5} & 54.31 & 65.60 & 25.34 & 61.76 & 1.191 & 0.185 \\
                        \addlinespace[0.3pt]
                        OmniCT-7B~\citep{lin2026omnict} & 41.80 & 45.40 & 14.00 & 57.00 & 0.943 & 0.147 \\
                        \addlinespace[0.3pt]
                        HealthGPT-Pro-8B~\citep{lin2025healthgpt} & 56.70 & 71.60 & 35.00 & 70.20 & 0.990 & 0.162 \\
                        \addlinespace[0.3pt]
                    ClinFusion-8B~\citep{yuan2026clinfusion} & 55.73 & 47.40 & 58.18 & 57.17 & 1.143 & 0.177 \\
                        \addlinespace[0.3pt]
                     GPT-5.4~\citep{openai_gpt54} & 63.93 & -- & 36.70 & -- & 0.763 & -- \\                                                                                                                                                                                                             
                    
                        \midrule
             
                        \multicolumn{7}{@{}l}{\textit{Decision VLMs}} \\

                        \addlinespace[0.3pt]
             
                         \addlinespace[0.3pt]
                        OmniMed-Jev~\citep{tang2026omnimed} & 59.90 & 43.80 & 54.20 & 61.80 & 0.558 & 0.123 \\
                        \addlinespace[0.3pt]
                        Intern-Decision-4B~\citep{zhu2026intern} & 63.30 & 64.60 & 53.70 & 58.60 & 0.321 & 0.106 \\
                        \addlinespace[0.3pt]
                        Jev-1.14~\citep{typesafe_jev_1_14} & 62.90 & 71.70 & 53.30 & 67.90 & 0.504 & 0.122 \\
                        \midrule
     
                        \addlinespace[0.3pt]                                                                                                                                                                                                                                                                   
            \textbf{Radev-0.8B (ours)} & \underline{69.96} & \underline{72.89} &  \underline{65.79} &  \underline{72.31} &  \underline{0.279} &  \underline{0.032} \\
            \addlinespace[0.3pt]
            \textbf{Radev-2B (ours)} & \textbf{72.72} & \textbf{77.61} & \textbf{69.29} & \textbf{75.16} & \textbf{0.234} & \textbf{0.028} \\
                        \bottomrule
                    \end{tabular}
                    \endgroup
                \end{table}
\noindent\textbf{Generalization on CT-Verdict-OOD.} 
Table~\ref{tab:model_comparison_2} evaluates model robustness under external distribution shifts. While most baseline models suffer on \textsc{Choice} accuracy and ordinal \textsc{Score} estimation, Radev-2B demonstrates strong cross-center generalization, consistently outperforming all competitors across every metric. Specifically, Radev-2B achieves 72.72\% Acc on \textsc{Noul} and 69.29\% Acc on \textsc{Choice}, surpassing the strongest closed-source models by ~9\% in binary and 16\% in multiple-choice accuracy. Radev-2B also attains the lowest ordinal KRPS and MAE,  with KRPS roughly $3.8\times$ below the strongest baseline (95\% study-bootstrap CI $[2.44, 7.20]$), showing that its severity distributions remain well-calibrated under external shift. 

\noindent\textbf{Comparison with zero-shot representation matching.} 
Table~\ref{tab:task_coverage} contrasts decision-native supervision with zero-shot representation matching. Embedding-similarity readouts support only disease-level discrimination: organ-level abnormality diagnosis, severity grading, triage, and laterality require structured verdict supervision that vector matching cannot provide, and their similarity scores are not calibrated probabilities. Radev supports all five task families with a single calibrated candidate readout and also achieves higher mAUC on both cohorts.


    \begin{table}[t]
        \centering
     \small
\caption{Comparison of Radev with SOTA methods on the diagnosis tasks from \textit{CT-Verdict-IID} and \textit{CT-Verdict-OOD}. DD denotes disease discrimination, OAD organ abnormality diagnosis, SG severity grading, AT acute triage, and LA laterality assessment. $\checkmark$ indicates supported and $\times$ indicates
unsupported.}
\label{tab:task_coverage}
\setlength{\tabcolsep}{15pt}
\begin{tabular}{@{}l*{9}{c}@{}}
\toprule
 \multirow{2}{*}{\textbf{Model}} & \multicolumn{5}{c}{\textbf{Clinical tasks}}
& \multicolumn{2}{c}{\textbf{IID}}
& \multicolumn{2}{c}{\textbf{OOD}} \\
\cmidrule(lr){2-6}
\cmidrule(lr){7-8}
\cmidrule(lr){9-10}
& DD & OAD & SG & AT & LA
& Acc & mAUC & Acc & mAUC \\
\midrule
Merlin-ZS
& $\checkmark$  & $\times$  &$\times$  & $\times$ & $\times$ 
 & 66.00 & 79.80 &  63.23     &  71.67   \\
\addlinespace[0.3pt]
RADAR
& $\checkmark$ & $\times$ & $\times$ & $\times$ & $\times$
& 66.73   & 80.10    & 60.50  & 74.71  \\
\midrule
\textbf{Radev-2B}
& $\checkmark$ & $\checkmark$ & $\checkmark$ & $\checkmark$ & $\checkmark$
& \textbf{90.08} & \textbf{87.79} & \textbf{72.72} & \textbf{77.61} \\
\bottomrule
\end{tabular}
\end{table}

\begin{figure}[h]
    \centering
    \includegraphics[width=0.79\linewidth]{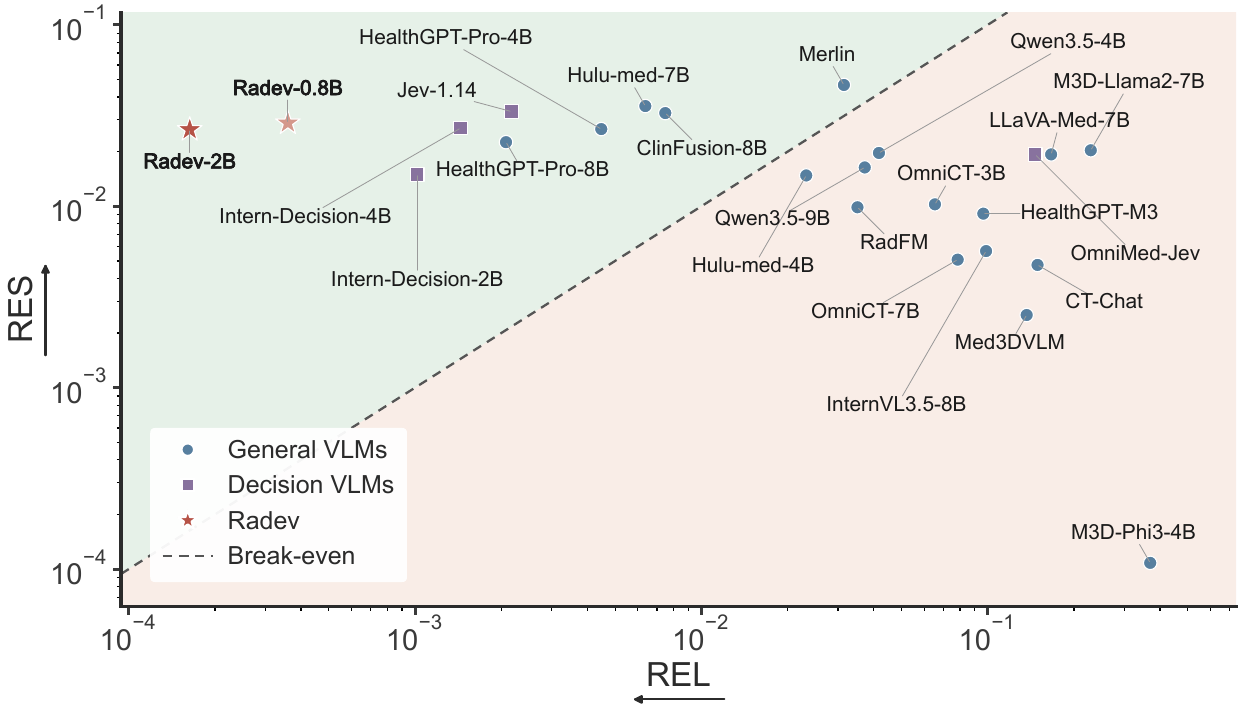}
   \vspace{1mm} 
    \caption{\textbf{Reliability resolution analysis.} Each point represents a model plotted by its reliability error (REL) and resolution (RES), computed from the binned Murphy decomposition of the top-label Brier score. The dashed line marks the $\mathrm{RES}=\mathrm{REL}$ boundary.}
    \label{fig:lik_murphy}
\end{figure}
    
\subsection{Confidence Analysis}
\noindent\textbf{How reliable are the confidence estimates?}
A model can predict the correct answer while assigning an inaccurate
confidence score. We therefore evaluate whether its confidence agrees
with how often its predictions are correct.
Figure~\ref{fig:lik_murphy} summarizes two aspects of confidence quality.
Reliability error (REL) measures the gap between average confidence and
accuracy within confidence groups, with lower values indicating better
calibration. Resolution (RES) measures how much accuracy differs across
these groups. Together, these terms assess whether assigning different
confidence scores to individual predictions improves on simply reporting
the model's overall accuracy for every prediction.
Under this evaluation, 14 of the 22 baselines perform worse than this
constant-confidence reference, whereas Radev achieves a positive Brier
skill score of 0.193 and a top-label ECE of 0.010. This reduces
calibration error by one to two orders of magnitude over the strongest baselines, demonstrating that Radev uniquely combines state-of-the-art
diagnostic accuracy with directly trustworthy confidence estimates.
\begin{figure}[h]
    \centering
    \includegraphics[width=0.79\linewidth]{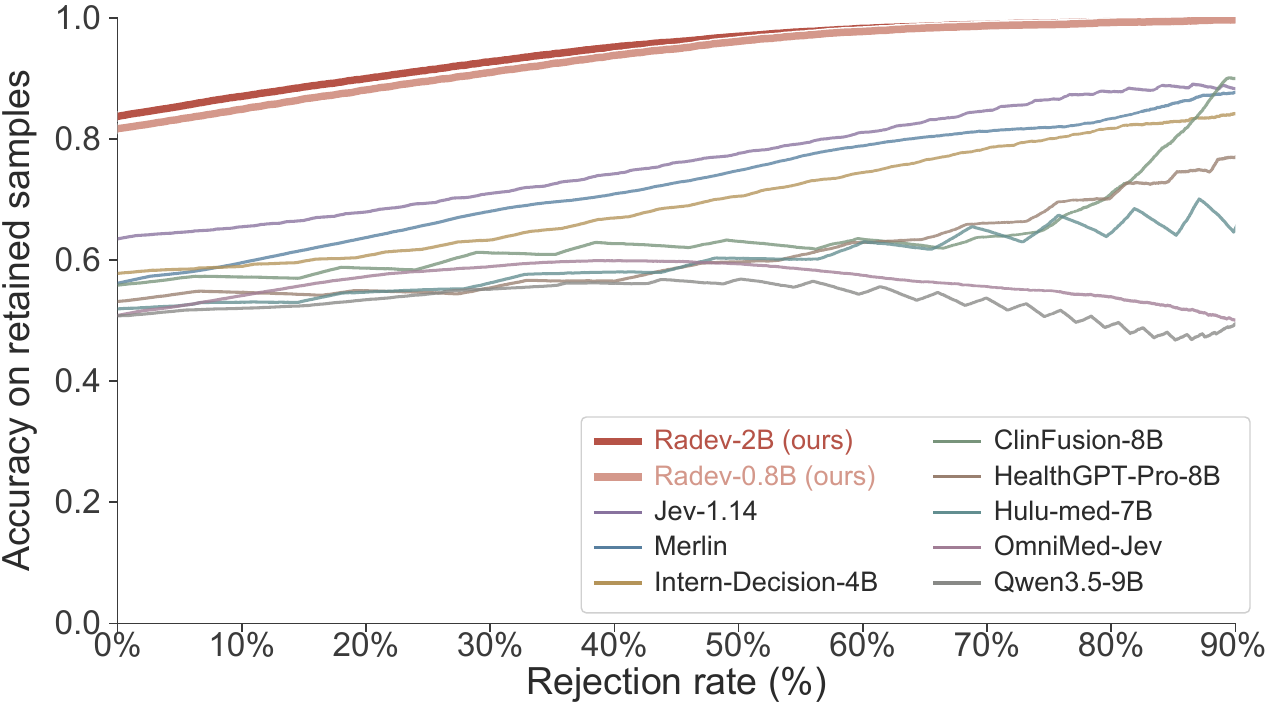}
 \vspace{1mm} 
    \caption{\textbf{Accuracy under confidence-based rejection.} Samples are rejected in ascending order of predictive confidence. The rejection rate is the percentage of all samples rejected, and accuracy is computed over the retained samples.}
    \label{fig:skill_rejection}
\end{figure}

\noindent\textbf{Selective prediction via confidence-based rejection.}
Figure~\ref{fig:skill_rejection} demonstrates how confidence-based rejection turns Radev's reported probabilities into an actionable clinical safety control. Deferring the least confident 20\%, 40\%, and 50\% of cases to human readers steadily raises Radev-2B's retained accuracy from 83.8\% to 90.0\%, 95.2\%, and 97.0\%. In contrast, no baseline matches this reliability at any deferral rate up to 90\%: the strongest baseline (Jev-1.14) attains only 77.6\% accuracy at 50\% deferral. Because Radev's accuracy--rejection curve is smooth, clinical workflows can select any operating point along the risk--workload frontier, allowing score autonomous verdicts and prioritize uncertain cases for clinician review.

\begin{table}[h]
        \centering
        \caption{Ablation study for the vision encoder architecture of Radev-2B after VFT on the \textit{CT-Verdict-IID}.}                                                                                                                                                                                     
        \label{tab:ablation_study}
        \begingroup
     \small
        \setlength{\tabcolsep}{1pt}
        \renewcommand{\arraystretch}{0.95}
        \begin{tabular*}{\columnwidth}{@{\extracolsep{\fill}}lccccccc@{}}
            \toprule
            \multirow{2}{*}{\textbf{Varian}t}
            & \multirow{2}{*}{\textbf{2D Slices}}
            & \multicolumn{2}{c}{\textbf{Noul}}
            & \multicolumn{2}{c}{\textbf{Choice}}
            & \multicolumn{2}{c}{\textbf{Score}} \\
            \cmidrule(lr){3-4}
            \cmidrule(lr){5-6}
            \cmidrule(lr){7-8}
            &
            & Acc$\uparrow$ & mAUC$\uparrow$
            & Acc$\uparrow$ & mAUC$\uparrow$
            & MAE$\downarrow$ & KRPS$\downarrow$ \\
            \midrule
             \addlinespace[0.3pt]
            2D encoder   & 16 & 83.20 & 69.07 & 63.63 & 62.49 & 0.43 & 0.07 \\
            \addlinespace[0.3pt]
            2D encoder   & 32 & 83.10 & 67.21 & 62.85 & 60.83 & 0.43 & 0.07 \\
            \addlinespace[0.3pt]
            3D encoder   & 0  & 88.11 & 83.10 & 67.52 & 77.40 & 0.41 & 0.07 \\
            \addlinespace[0.3pt]
            token concat & 32  & 88.69   & 84.19    & 68.29  & 77.03  & 0.41        & 0.07 \\
     \midrule
            \addlinespace[0.3pt]
               \addlinespace[0.3pt]
                w/o grounding & 32 & 89.33 & 85.95 & 69.51 & 80.36 & 0.39 & \textbf{0.06} \\
               w/o organ emb.  & 32 & 88.36 & 85.24 & 67.61 & 77.38 & 0.40 & 0.06 \\
                w/o slot $n{+}1$  & 32 & 86.93 & 82.53 & 67.81 & 75.25 & 0.40 & 0.06 \\
             \midrule
            \addlinespace[0.3pt]
            organ-grounded visual        & 4  & 88.79 & 85.53 & 69.10 & 77.99 & 0.39 & 0.06 \\
            \addlinespace[0.3pt]
            organ-grounded visual        & 8  & 88.66 & 84.91 & 68.67 & 78.92 & 0.39 & 0.06 \\
            \addlinespace[0.3pt]
            organ-grounded visual       & 16 & 89.14 & 86.24 & 69.63 & 80.27 & \textbf{0.39} & 0.06 \\
            \addlinespace[0.3pt]
            organ-grounded visual        & 32 & \textbf{89.91} & \textbf{87.89} & \textbf{70.30} & \textbf{81.41} & 0.39 & 0.06 \\
            \bottomrule
        \end{tabular*}
        \endgroup
    \end{table}
      \begin{table}[h]
            \centering
            \caption{Comparison results of Radev-2B-VFT and baselines after VFT in the same training configurations on the \textit{CT-Verdict-IID}.}
            \label{tab:backbone_comparison}
            \begingroup
         \small
            \setlength{\tabcolsep}{1pt}
            \renewcommand{\arraystretch}{0.95}
            \begin{tabular*}{\columnwidth}{@{\extracolsep{\fill}}lcccccc@{}}
                \toprule
                \multirow{2}{*}{\textbf{Model}}
 
                & \multicolumn{2}{c}{\textbf{Noul}}
                & \multicolumn{2}{c}{\textbf{Choice}}
                & \multicolumn{2}{c}{\textbf{Score}} \\
                \cmidrule(lr){2-3}\cmidrule(lr){4-5}\cmidrule(lr){6-7}
                &  Acc$\uparrow$ & mAUC$\uparrow$
                & Acc$\uparrow$ & mAUC$\uparrow$
                & MAE$\downarrow$ & KRPS$\downarrow$ \\
                \midrule
                \addlinespace[0.3pt]
                Qwen3.5-2B~\citep{qwen3.5}  & 83.10 & 67.21 & 62.85 & 60.83 & 0.43 & 0.07 \\
                \addlinespace[0.3pt]
                Qwen3.5-4B~\citep{qwen3.5}  & 82.96 & 67.45 & 62.93 & 61.42 & 0.43 & 0.07 \\
                \addlinespace[0.3pt]
                ClinFusion-8B~\citep{yuan2026clinfusion}  & 80.36 & 51.23 & 59.34 & 50.20 & 0.46 & 0.07 \\
                \addlinespace[0.3pt]
             \textbf{Radev-2B-VFT (ours)} &\textbf{89.91} & \textbf{87.89} & \textbf{70.30} & \textbf{81.41} & \textbf{0.39} & \textbf{0.06} \\
                \bottomrule
            \end{tabular*}
            \endgroup
        \end{table}
\subsection{Ablation Studies}

\noindent\textbf{Vision encoder architecture.}
Table~\ref{tab:ablation_study} ablates the organ-grounded visual encoder on
\textit{CT-Verdict-IID} after VFT. Pure 2D slice encoders struggle to capture
volumetric continuity. Conversely, a pure 3D encoder substantially boosts volumetric
discrimination, and naive 2D--3D token concatenation yields
only marginal gains. Replacing naive fusion with our design achieves the best overall performance across all tasks. Component-wise ablations confirm the role of
each design in our organ-ledger representation: removing 3D spatial box grounding, learned organ identity embeddings, or
the global context slot consistently hurts diagnostic
discrimination. Finally, scaling the 2D stream
to 32 slices under our design yields monotonic gains in mAUC across both
\textsc{noul} and \textsc{choice}, demonstrating that organ-gated cross-attention
effectively harvests fine-grained slice cues without suffering from token dilution. To isolate architectural gains from training data effects,
Table~\ref{tab:backbone_comparison} compares Radev-2B-VFT with baselines fine-tuned under the identical VFT configuration on \textit{CT-Verdict}.
Radev-2B-VFT substantially outperforms these larger
models across all metrics, confirming
that the designed architecture is the primary driver of our performance lead.

   \begin{table}[H]
        \centering
        \caption{Ablation study for the training stages of Radev-2B on the \textit{CT-Verdict-IID}, evaluated at both the accuracy (top) and calibration (bottom, $\times 100$) levels.}
        \label{tab:stage_ablation}
        \begingroup
     \small
        \setlength{\tabcolsep}{16pt}
        \renewcommand{\arraystretch}{0.95}
        \begin{tabular}{@{}lcccccccc@{}}
            \toprule
            \multirow{2}{*}{\textbf{Align}} & \multirow{2}{*}{\textbf{VFT}} & \multirow{2}{*}{\textbf{CCT}} & \multicolumn{2}{c}{\textbf{Noul}} & \multicolumn{2}{c}{\textbf{Choice}} & \multicolumn{2}{c}{\textbf{Score}} \\
            \cmidrule(lr){4-5} \cmidrule(lr){6-7} \cmidrule(lr){8-9}
             & & & Acc$\uparrow$ & mAUC$\uparrow$ & Acc$\uparrow$ & mAUC$\uparrow$ & MAE$\downarrow$ & KRPS$\downarrow$ \\
            \midrule
            $\checkmark$ & $\times$ & $\times$ & 86.93 & 80.78 & 66.15 & 72.27 & 0.46 & 0.07 \\
            \addlinespace[0.3pt]
            $\times$ & $\checkmark$ & $\times$ & 88.93 & 85.50 & 68.26 & 77.65 & 0.40 & 0.06 \\
            \addlinespace[0.3pt]
            $\checkmark$ & $\checkmark$ & $\times$ & 89.91 & \textbf{87.89} & 70.30 & 81.41 & \textbf{0.39} & 0.06 \\
            \addlinespace[0.3pt]
            $\checkmark$ & $\checkmark$ & $\checkmark$ & \textbf{90.08} & 87.79 & \textbf{71.33} & \textbf{81.91} & 0.39 & \textbf{0.06} \\
            \bottomrule
        \end{tabular}
        \endgroup
        \begingroup
     \small
        \setlength{\tabcolsep}{17pt}
        \renewcommand{\arraystretch}{0.95}
        \begin{tabular}{@{}lcccccccc@{}}
            \toprule
           \multirow{2}{*}{\textbf{Align}} & \multirow{2}{*}{\textbf{VFT}} & \multirow{2}{*}{\textbf{CCT}} & \multicolumn{2}{c}{\textbf{Noul}} & \multicolumn{2}{c}{\textbf{Choice}} & \multicolumn{2}{c}{\textbf{Score}} \\
            \cmidrule(lr){4-5} \cmidrule(lr){6-7} \cmidrule(lr){8-9}
             & & & Brier$\downarrow$ & ECE$\downarrow$ & Brier$\downarrow$ & ECE$\downarrow$ & Brier$\downarrow$ & ECE$\downarrow$ \\
            \midrule
            $\checkmark$ & $\times$ & $\times$ & 19.05 & 1.96 & 47.70 & 2.88 & 50.39 & 5.63 \\
            \addlinespace[0.3pt]
            $\times$ & $\checkmark$ & $\times$ & 16.47 & 1.10 & 44.33 & 2.57 & 43.27 & 6.89 \\
            \addlinespace[0.3pt]
            $\checkmark$ & $\checkmark$ & $\times$ & 14.96 & 2.00 & 41.78 & 2.98 & 43.34 & 9.06 \\
            \addlinespace[0.3pt]
            $\checkmark$ & $\checkmark$ & $\checkmark$ & \textbf{14.57} & \textbf{0.57} & \textbf{40.61} & \textbf{2.09} & \textbf{41.03} & \textbf{2.04} \\
            \bottomrule
        \end{tabular}
        \endgroup
    \end{table}

\noindent\textbf{Training strategy.}
Table~\ref{tab:stage_ablation} examines the progression across organ-ledger alignment, VFT, and CCT. Combining alignment with VFT substantially improves
diagnostic accuracy over either stage alone. Adding CCT not only preserves or
further improves diagnostic performance, but also sharply reduces Brier scores and slashes ECE on each task. Table~\ref{tab:cct_abl_calib} further decomposes how CCT achieves this calibration gain
via Murphy decomposition (reliability $\text{REL}$ and resolution
$\text{RES}$). Introducing multi-view counterfactuals and joint task supervision
across the factual, target-ablated, and control-ablated views delivers the primary calibration leap, cutting ECE to $0.60 / 2.26 / 2.20$
while maintaining discrimination. This empirically validates our
mechanism in Sec.~\ref{sec:cct}: supervising visually indistinguishable target-ablated
views with their original one-hot gold labels naturally pulls overconfident predictions
toward the conditional empirical prior. Finally, incorporating the evidence-blind view and the evidence-dependence hinge $\mathcal{L}_{\mathrm{blind}}$ discourages evidence-independent confidence, achieving the lowest REL and ECE.

\begin{table}[H]
    \centering
    \caption{CCT ablation on the \textit{CT-Verdict-IID} to evaluate calibration (Murphy decomposition, $\times100$).}
    \label{tab:cct_abl_calib}
    \setlength{\tabcolsep}{9.0pt}
 \small
    \begin{tabular}{l ccc ccc ccc}
    \toprule
     \multirow{2}{*}{\textbf{Variant}}   & \multicolumn{3}{c}{\textbf{Noul}} & \multicolumn{3}{c}{\textbf{Choice}} & \multicolumn{3}{c}{\textbf{Score}} \\
    \cmidrule(lr){2-4}\cmidrule(lr){5-7}\cmidrule(lr){8-10}
  & REL$\downarrow$ & RES$\uparrow$ & ECE$\downarrow$
            & REL$\downarrow$ & RES$\uparrow$ & ECE$\downarrow$
            & REL$\downarrow$ & RES$\uparrow$ & ECE$\downarrow$ \\
    \midrule
    continue VFT                & 0.09 & 33.52 & 0.99 & 0.79 & \textbf{23.65} & 3.13 & 0.71 & \textbf{18.46} & 4.17 \\
    + multi-view             & 0.05 & \textbf{33.54} & 1.15 & 0.66 & 23.63 & 3.45 & 0.45 & 18.17 & 3.07 \\
    + all-view sup. & \textbf{0.02} & 33.40 & 0.60 & 0.56 & 23.56 & 2.26 & 0.38 & 18.34 & 2.20 \\
   + blind view \& hinge  & 0.02 & 33.39 & \textbf{0.57} & \textbf{0.55} & 23.58 & \textbf{2.09} & \textbf{0.35} & 18.25 & \textbf{2.04} \\
    \bottomrule
    \end{tabular}
    \end{table}   
 
\section{Conclusion}
\label{sec:conclusion}

We presented \textbf{Radev}, a decision-native multimodal framework for structured 3D
abdominal CT interpretation, alongside the \textbf{CT-Verdict} benchmark. By coupling a
reusable 3D organ ledger with question-bound 2D slice evidence, Radev predicts
closed-set answer distributions in a single forward pass without autoregressive
decoding, while CCT anchors confidence in
anatomical evidence by contrasting intact, ablated, and evidence-blind views. Across
both \textit{CT-Verdict-IID} and external \textit{CT-Verdict-OOD} evaluations, Radev
achieves state-of-the-art diagnostic accuracy and calibration skill, delivering directly trustworthy probabilities for selective
prediction and clinical abstention. In
future work, we will extend the organ-grounded ledger and counterfactual calibration
framework to other volumetric imaging modalities, particularly multi-sequence MRI, as
well as broader anatomical regions.


\bibliographystyle{assets/plainnat}
\bibliography{paper}

\newpage
\beginappendix
\setcounter{page}{1}

\appendix  

\setcounter{figure}{0}
\setcounter{table}{0}
\setcounter{equation}{0}

\renewcommand{\thefigure}{A\arabic{figure}}
\renewcommand{\thetable}{A\arabic{table}}
\renewcommand{\theequation}{A\arabic{equation}}
 
\section*{Supplementary Outline}

\begin{itemize}[leftmargin=*, nosep]
    \item \textbf{Section~\ref{sec:supp-bench}}: Additional CT-Verdict Details
    \begin{itemize}[nosep]
        \item Section~\ref{sec:supp-stats}: Dataset Statistics (Anatomical, Clinical Tasks, and Disease Distributions)
        \item Section~\ref{sec:supp-verification}: Verification and Quality Control (Multi-reader LLM Audit, Human Audit, Text-only Audit)
        \item Details for Evaluation Protocol (Classification, Ordinal, and Calibration Metrics)
    \end{itemize}

    \item \textbf{Section~\ref{sec:supp-method}: Additional Method Details}
    \begin{itemize}[nosep]
        \item Section~\ref{sec:supp-organ-stratified}: Organ-stratified Curriculum Sampling Strategy.
        \item Section~\ref{sec:supp-proof}: Population-Level Analysis of Counterfactual Confidence Tempering (Proposition 1 \& Proof)
        \item Section~\ref{sec:supp-fusion}: 2D-3D Feature Fusion in Vision Encoder (Spatial Membership Rule)
    \end{itemize}
\end{itemize}
 \section{Additional CT-Verdict Details}
 \label{sec:supp-bench}
 \begin{figure*}[ht]
    \centering
    \includegraphics[width=\textwidth]{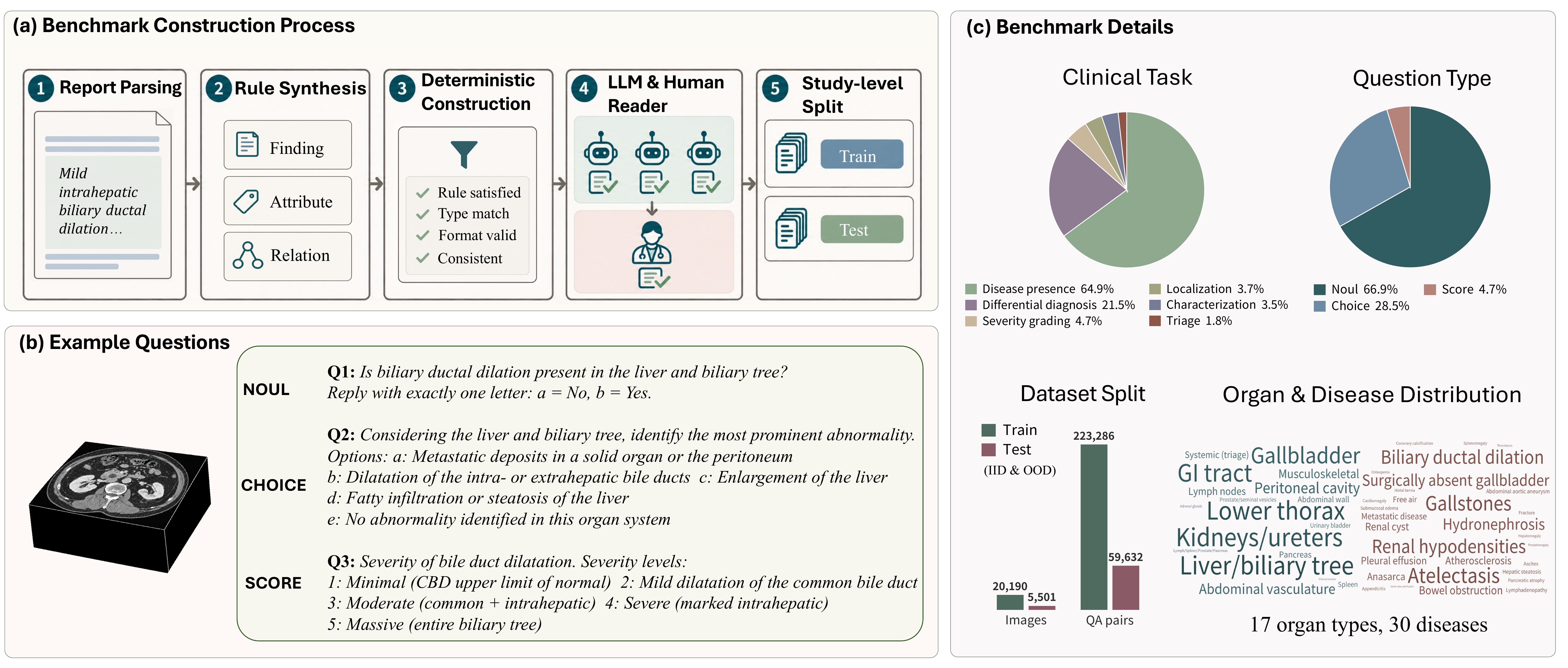}
    \vspace{1mm} 
 \caption{\textbf{Overview of \textit{CT-Verdict.}}
    \textbf{(a)} Benchmark construction process: radiology reports are parsed into
    structured findings; synthesis rules over presence,
    attributes, and relations turn each finding into candidate questions; questions are generated deterministically and their labels verified by three LLM
    readers with human spot-checking; finally, a
    study-level split separates training from test.
    \textbf{(b)} Examples of the three question types: binary presence (noul),
    multi-way choice, and ordinal severity (score).
    \textbf{(c)} Details of dataset: distributions of question types and clinical
    tasks, dataset splits, and organ/disease coverage.}
    \label{fig:benchmark}
\end{figure*}
\label{sec:appendix_ct_verdict}
This section summarizes the more details that supplement the benchmark description in the main text. Overview of the benchmark can be seen in Figure~\ref{fig:benchmark}.

\subsection{Dataset Statistics}
\label{sec:supp-stats}

\paragraph{Anatomical coverage.}
\textit{CT-Verdict} spans 17 anatomical organ and region categories, with detailed
per-split statistics provided in Table~\ref{tab:supp-organs}. The training (223{,}286
questions) and \textit{CT-Verdict-IID} test (56{,}825 questions) splits exhibit nearly
identical anatomical distributions, led by the liver/biliary tree, lower thorax,
kidneys/ureters, gastrointestinal tract, and gallbladder. In contrast, the external
\textit{CT-Verdict-OOD} test split focuses on core abdominal anatomy (led by
kidneys/ureters, peritoneal cavity, and liver/biliary tree) while omitting
lower-thorax, systemic/triage, abdominal-wall, bladder, and uterus categories.

\paragraph{Clinical tasks and disease distribution.}
Tables~\ref{tab:supp-capabilities} and~\ref{tab:supp-diseases} report the distributions
across the five major clinical capabilities (plus a residual ``Other'' category of
fine-grained attribute questions) and the 30 annotated diseases/findings for the
training, IID test, and OOD test splits. The training and IID test splits match closely
across both task capabilities and individual findings, following a natural long-tailed
clinical distribution. Conversely, \textit{CT-Verdict-OOD} concentrates on disease
discrimination and organ abnormality identification (without acute triage or laterality
items), omits six thoracic-adjacent or cardiovascular findings, and re-weights toward
targeted abdominal pathologies such as lymphadenopathy, ascites, splenomegaly, and
renal cysts.

\begin{table}[h]
\centering
\small
\setlength{\tabcolsep}{2.5pt}
\renewcommand{\arraystretch}{0.95}
\caption{Distribution of major anatomical organ/region categories across the training, IID test, and OOD test splits. Values are $n$ (\%).}
\label{tab:supp-organs}
\begin{tabularx}{\columnwidth}{@{} >{\raggedright\arraybackslash}X r r r @{}}
\toprule
\textbf{Organ / Region} & \textbf{Train} & \textbf{IID} & \textbf{OOD} \\
\midrule
Liver / biliary tree    & 33{,}483 (15.0\%) & 8{,}568 (15.1\%) & 248 (12.1\%) \\
Lower thorax            & 33{,}427 (15.0\%) & 8{,}414 (14.8\%) & 0 (0.0\%)    \\
Kidneys / ureters       & 32{,}937 (14.8\%) & 8{,}308 (14.6\%) & 381 (18.5\%) \\
Gastrointestinal tract  & 30{,}443 (13.6\%) & 7{,}540 (13.3\%) & 197 (9.6\%)  \\
Gallbladder             & 24{,}295 (10.9\%) & 6{,}041 (10.6\%) & 143 (7.0\%)  \\
Peritoneal cavity       & 21{,}454 (9.6\%)  & 5{,}494 (9.7\%)  & 316 (15.4\%) \\
Vasculature             & 18{,}089 (8.1\%)  & 4{,}740 (8.3\%)  & 75 (3.6\%)   \\
Remaining categories    & 29{,}158 (13.1\%) & 7{,}720 (13.6\%) & 695 (33.8\%) \\
\bottomrule
\end{tabularx}
\end{table}

\begin{table}[h]
\centering
\small
\setlength{\tabcolsep}{2.5pt}
\renewcommand{\arraystretch}{0.95}
\caption{Distribution of clinical task capabilities across the training, IID test, and OOD test splits. Values are $n$ (\%). Each question carries one capability label; ``Other'' groups fine-grained attribute questions, and percentages use each split's total.}
\label{tab:supp-capabilities}
\begin{tabularx}{\columnwidth}{@{} >{\raggedright\arraybackslash}X r r r @{}}
\toprule
\textbf{Clinical capability} & \textbf{Train} & \textbf{IID} & \textbf{OOD} \\
\midrule
Disease discrimination & 138{,}828 (62.2\%) & 35{,}265 (62.1\%) & 1{,}194 (58.1\%) \\
Organ abnormality      & 47{,}836 (21.4\%)  & 12{,}178 (21.4\%) & 769 (37.4\%)     \\
Severity grading       & 10{,}521 (4.7\%)   & 2{,}690 (4.7\%)   & 59 (2.9\%)       \\
Acute triage           & 3{,}963 (1.8\%)    & 1{,}034 (1.8\%)   & 0 (0.0\%)        \\
Laterality assessment  & 984 (0.4\%)        & 259 (0.5\%)       & 0 (0.0\%)        \\
Other                  & 21{,}154 (9.5\%)   & 5{,}399 (9.5\%)   & 33 (1.6\%)       \\
\bottomrule
\end{tabularx}
\end{table}

\begin{table}[h]
\centering
\small
\setlength{\tabcolsep}{2.5pt}
\renewcommand{\arraystretch}{0.92}
\caption{Question counts and shares across the 30 annotated diseases/findings in the training (150{,}333 disease-tagged items), IID test (38{,}214 items), and OOD test (1{,}253 items) splits, counting only disease-tagged questions. Values are $n$ (\%).}
\label{tab:supp-diseases}
\begin{tabularx}{\columnwidth}{@{} >{\raggedright\arraybackslash}X r r r @{}}
\toprule
\textbf{Disease / Finding} & \textbf{Train} & \textbf{IID} & \textbf{OOD} \\
\midrule
Atelectasis                   & 13{,}687 (9.1\%) & 3{,}434 (9.0\%) & 0 (0.0\%)    \\
Gallstones                    & 12{,}096 (8.0\%) & 2{,}996 (7.8\%) & 59 (4.7\%)   \\
Renal hypodensities           & 9{,}594 (6.4\%)  & 2{,}370 (6.2\%) & 110 (8.8\%)  \\
Biliary ductal dilation       & 9{,}443 (6.3\%)  & 2{,}492 (6.5\%) & 82 (6.5\%)   \\
Surgically absent gallbladder & 9{,}392 (6.2\%)  & 2{,}298 (6.0\%) & 33 (2.6\%)   \\
Bowel obstruction             & 9{,}063 (6.0\%)  & 2{,}240 (5.9\%) & 35 (2.8\%)   \\
Hydronephrosis                & 8{,}990 (6.0\%)  & 2{,}368 (6.2\%) & 78 (6.2\%)   \\
Renal cyst                    & 7{,}418 (4.9\%)  & 1{,}867 (4.9\%) & 123 (9.8\%)  \\
Atherosclerosis               & 7{,}406 (4.9\%)  & 1{,}953 (5.1\%) & 21 (1.7\%)   \\
Pleural effusion              & 7{,}369 (4.9\%)  & 1{,}828 (4.8\%) & 0 (0.0\%)    \\
Anasarca                      & 7{,}128 (4.7\%)  & 1{,}781 (4.7\%) & 0 (0.0\%)    \\
Metastatic disease            & 6{,}934 (4.6\%)  & 1{,}732 (4.5\%) & 11 (0.9\%)   \\
Free air                      & 4{,}842 (3.2\%)  & 1{,}237 (3.2\%) & 17 (1.4\%)   \\
Abdominal aortic aneurysm     & 4{,}549 (3.0\%)  & 1{,}250 (3.3\%) & 17 (1.4\%)   \\
Hepatic steatosis             & 4{,}378 (2.9\%)  & 1{,}099 (2.9\%) & 34 (2.7\%)   \\
Lymphadenopathy               & 3{,}829 (2.5\%)  & 1{,}051 (2.8\%) & 155 (12.4\%) \\
Ascites                       & 3{,}323 (2.2\%)  & 882 (2.3\%)     & 102 (8.1\%)  \\
Submucosal edema              & 2{,}746 (1.8\%)  & 711 (1.9\%)     & 57 (4.5\%)   \\
Pancreatic atrophy            & 2{,}610 (1.7\%)  & 696 (1.8\%)     & 60 (4.8\%)   \\
Fracture                      & 2{,}573 (1.7\%)  & 674 (1.8\%)     & 17 (1.4\%)   \\
Appendicitis                  & 2{,}400 (1.6\%)  & 556 (1.5\%)     & 22 (1.8\%)   \\
Coronary calcification        & 1{,}766 (1.2\%)  & 445 (1.2\%)     & 0 (0.0\%)    \\
Cardiomegaly                  & 1{,}709 (1.1\%)  & 412 (1.1\%)     & 0 (0.0\%)    \\
Hiatal hernia                 & 1{,}487 (1.0\%)  & 363 (0.9\%)     & 10 (0.8\%)   \\
Splenomegaly                  & 1{,}439 (1.0\%)  & 434 (1.1\%)     & 96 (7.7\%)   \\
Osteopenia                    & 1{,}323 (0.9\%)  & 371 (1.0\%)     & 9 (0.7\%)    \\
Hepatomegaly                  & 1{,}249 (0.8\%)  & 320 (0.8\%)     & 44 (3.5\%)   \\
Thrombosis                    & 723 (0.5\%)      & 155 (0.4\%)     & 20 (1.6\%)   \\
Prostatomegaly                & 666 (0.4\%)      & 143 (0.4\%)     & 41 (3.3\%)   \\
Aortic valve calcification    & 201 (0.1\%)      & 56 (0.1\%)      & 0 (0.0\%)    \\
\bottomrule
\end{tabularx}
\end{table}

\subsection{Verification and Quality Control}
\label{sec:supp-verification}

\paragraph{Multi-reader verification and filtering.}
To ensure high label fidelity, candidate questions generated by our rule-based pipeline
undergo a strict multi-reader verification gate before release. As shown in
Table~\ref{tab:supp-reader}, the two base LLM readers achieve strong consensus on
audited report units, with an inter-reader agreement of 97.8\%  on
\textit{CT-Verdict-IID} and 96.7\% on \textit{CT-Verdict-OOD}. Items receiving split
votes from the base readers are routed to a third adjudicating reader, and any item
failing to reach a two-of-three majority is discarded rather than relabeled. Following
verification, release-level filtering removes template-only statements and balances
option distributions, yielding the final released corpus of 280{,}111 questions for
\textit{CT-Verdict-IID} (223{,}286 training/validation and 56{,}825 test).

\begin{table}[h]
\centering
\small
\setlength{\tabcolsep}{5pt}
\renewcommand{\arraystretch}{1.0}
\caption{Multi-reader verification statistics on audited finding-presence units.}
\label{tab:supp-reader}
\begin{tabular*}{\columnwidth}{@{\extracolsep{\fill}}lcc@{}}
\toprule
\textbf{Metric} & \textbf{CT-Verdict-IID} & \textbf{CT-Verdict-OOD} \\
\midrule
Audited report units    & 82{,}836          & 613          \\
Base-reader consensus   & 79{,}494 (96.0\%) & 567 (92.5\%) \\
Adjudicated units       & 1{,}854           & 19           \\
Adjudication retained   & 1{,}564           & 9            \\
Inter-reader agreement  & 97.8\%            & 96.7\%       \\
\bottomrule
\end{tabular*}

\vspace{2pt}
\parbox{\columnwidth}{\small
A report unit is one disease-presence assertion for a study, tied to its source sentence and rule-derived polarity label; the two base readers independently judge each unit.
}
\end{table}

\paragraph{Human audit and split integrity.}
To validate the final released labels, we manually audited random samples of 400 items
from \textit{CT-Verdict-IID} and 200 items from \textit{CT-Verdict-OOD} against their
source radiology reports. In the IID sample, all 400 items were verified as clinically
acceptable with zero factual contradictions against the source text. In the OOD sample,
following a secondary curation pass, all 200 sampled items were likewise confirmed to be
factually aligned with the source reports, achieving 100\% audit accuracy across both
released benchmarks. Finally, we verified that no study identifier,
report text, or image file is shared between the training and test splits, and that the
OOD cohort is used strictly for zero-shot evaluation.

\paragraph{Text-only audit}

To verify that \textit{CT-Verdict} genuinely requires visual interpretation rather than
question-surface shortcuts, and that Radev's high performance does not simply
stem from memorizing training-set linguistic priors, we conduct a strict no-image audit
on the full \textit{CT-Verdict-IID} test split. At inference, we
completely zero out the visual pathway, leaving the question
text and readout pipeline unchanged. Withholding visual input triggers a sharp collapse across all discriminative and
clinical-grading metrics. On binary presence and multiple-choice questions,
probability-based mAUC plunges by $22.67$ and $22.63$ points (from $87.79\%$ to
$65.12\%$ and from $81.91\%$ to $59.28\%$, respectively), while balanced accuracy
(mBAcc) drops from $71.79\%$ to near-chance ($53.32\%$). On ordinal severity grading,
removing the image increases MAE from $0.392$ to $0.470$ and cuts rank agreement (QWK)
in half, from $0.537$ to $0.264$. The degradation is even more pronounced on tasks that
cannot be inferred from routine reporting statistics, such as acute triage, where
accuracy drops by $45.36$ points ($82.88\%$ to $37.52\%$). Running the same no-image
control on a frontier closed-source model (GPT-5.4) yields an analogous breakdown, with
its ordinal severity QWK collapsing from $0.462$ to $0.105$ (and MAE rising from
$0.748$ to $0.983$). Together, these results confirm that \textit{CT-Verdict} also cannot be
solved via text priors alone and that Radev's performance is driven by genuine
anatomical visual grounding.

\subsection{Details for Evaluation Protocol}
\paragraph{Classification Metrics (Binary \& Multiple-Choice).} 
We report per-question exact-match accuracy (\textbf{Acc}), macro-averaged balanced accuracy (\textbf{mBAcc}), and macro-averaged area under the ROC curve (\textbf{mAUC}). Both mBAcc and mAUC are first computed one-vs-rest within each disease category's questions: mBAcc as the mean of sensitivity and specificity, and mAUC using the predicted probability (the ``Yes'' probability for binary, and the option probability for multiple-choice) against the one-vs-rest gold label. These scores are then macro-averaged across categories, specifically 30 clinical findings for binary and 26 option diseases for multiple-choice, to weight rare and common findings equally. The overall mBAcc averages the two branches per finding before macro-averaging. In contrast, mAUC is reported separately per branch because the two probability scales are incommensurate.

\paragraph{Ordinal Severity Metrics.} 
Ordinal questions are evaluated using mean absolute error (\textbf{MAE}), normalized ranked probability score (\textbf{KRPS})~\cite{epstein1969scoring}, and quadratic weighted kappa (\textbf{QWK})~\cite{cohen1968weighted}. 
KRPS compares the predicted and gold cumulative level distributions,
\begin{equation}
\mathrm{KRPS}_i \;=\; \frac{1}{K-1}\sum_{k=1}^{K-1}\bigl(\hat F_i(k)-F_i(k)\bigr)^2,
\end{equation}
penalizing probability mass on every incorrect level rather than only the argmax. It is omitted for unordered questions, which lack cumulative structure. 
QWK discounts quadratic-weighted agreement by the chance agreement expected from marginal distributions, assigning a score of zero to a prevalence-only predictor.

\paragraph{Calibration and Murphy Decomposition.} 
Calibration is assessed at two complementary levels using ten equal-width bins (dropping empty bins):
\begin{itemize}
    \item \textbf{Decision level:} The top-label expected calibration error (\textbf{ECE})~\cite{guo2017calibration} evaluates the confidence of the argmax prediction:
    \begin{equation}
    \mathrm{ECE} \;=\; \sum_b \frac{n_b}{N}\,\bigl|\mathrm{acc}_b-\mathrm{conf}_b\bigr|,
    \end{equation}
    where $n_b$ is the size of bin $b$, and $\mathrm{acc}_b$ and $\mathrm{conf}_b$ are its accuracy and mean top-label confidence.
    \item \textbf{Distribution level:} The multiclass Brier score (\textbf{BS})~\cite{glenn1950verification} is a proper scoring rule assessing the entire candidate distribution:
    \begin{equation}
    \mathrm{BS} \;=\; \frac{1}{N}\sum_i\sum_k \bigl(p_{ik}-y_{ik}\bigr)^2,
    \end{equation}
    with $p_{ik}$ the predicted probability of candidate $k$ and $y_{ik}$ its one-hot gold indicator. To disentangle miscalibration from discrimination, we apply the Murphy decomposition~\cite{murphy1973new} to its binned form:
    \begin{equation}
    \mathrm{BS} \;=\; \mathrm{REL}-\mathrm{RES}+\mathrm{UNC},
    \end{equation}
    where the three terms are
    \begin{equation}
    \begin{aligned}
    \mathrm{REL} &= \sum_b w_b\,(\bar p_b-\bar y_b)^2,\\
    \mathrm{RES} &= \sum_b w_b\,(\bar y_b-\bar y)^2,\\
    \mathrm{UNC} &= \bar y\,(1-\bar y).
    \end{aligned}
    \end{equation}
    \textbf{REL} measures pure miscalibration (squared confidence--accuracy gap, lower is better), \textbf{RES} measures discrimination (spread of bin accuracies around the base rate, higher is better), and \textbf{UNC} is the data-intrinsic uncertainty independent of the model.
\end{itemize}
The decomposition identity holds exactly for the binned score. The raw row-wise BS reported alongside differs from it only by a within-bin variance term. For multi-candidate questions, the decomposition is applied one-vs-rest per candidate, matching the multiclass BS projection whereas ECE targets the top-label decision. It is then summed and weighted by candidate-count buckets to yield the reported REL and RES. A model outperforms a constant-prevalence baseline if and only if $\mathrm{RES}>\mathrm{REL}$ (positive Brier skill score), which defines the break-even boundary in Fig.~\ref{fig:lik_murphy}.

 \section{Additional Method Details}
\label{sec:supp-method}
\subsection{Organ-Stratified Curriculum of CCT}
\label{sec:supp-organ-stratified}
Because multi-view calibration acts directly on ablated slots, uniform sampling would
under-calibrate rare anatomy. We therefore sample each of the 17 organ groups $g$ in
CT-Verdict of size $n_g$ with probability
\begin{equation}
  \pi_g\;\propto\; n_g^{\,\tau},\qquad \tau=0.5,
  \label{eq:cct-sampler}
\end{equation}
interpolating between instance- and organ-uniform distributions. Group quotas are
floored at $0.5\%$ of the budget, capped at eight passes per group, and interleaved at
the study-run level so that even a short CCT budget (Sec.~\ref{sec:impl}) maintains
balanced calibration across all organs.

\subsection{Analysis of Counterfactual Confidence Tempering}
\label{sec:supp-proof}

We provide a population-level analysis motivating how supervising counterfactual organ
views under Eq.~\eqref{eq:loss-kinds} and Eq.~\eqref{eq:cct-blind} couples predictive
confidence to anatomical evidence rather than shortcut cues; in practice, the magnitude
of this effect depends on finite-sample optimization and model smoothness, and the
zero-information condition for control ablation is an idealization that holds only
approximately, since unbound organs may carry residual information about the verdict.

\paragraph{Proposition 1 (Posterior tempering under target ablation).}
\emph{Let $Y \in \mathcal{Y}(q)$ denote the reference verdict for question $q$.
Partition the factual organ ledger as $\mathbf{E}=(\mathbf{E}_T,\mathbf{E}_R)$, where
$\mathbf{E}_T$ collects all components removed by target ablation: the ledger tokens of
the question-bound target organs $\mathcal{B}(q)$, their bound 2D evidence tokens, and
the global slot $n{+}1$, while $\mathbf{E}_R$ denotes the remaining unbound organs and
their evidence.}

\emph{Let $\tilde{\mathbf{E}}$ denote the target-ablated ledger obtained by replacing
the entire block $\mathbf{E}_T$ with the shared placeholder sequence $\mathbf{a}$ (each
affected evidence token reducing to its intervened ledger token,
Sec.~\ref{sec:cct}) while retaining $\mathbf{E}_R$, and let $\hat{\mathbf{E}}$ denote
the control-ablated ledger obtained by replacing one irrelevant organ
$\mathbf{E}_c \in \mathbf{E}_R$ (its ledger token and bound evidence) with $\mathbf{a}$
while leaving $\mathbf{E}_T$ unchanged.}

\emph{For each view $\mathbf{V} \in \{\mathbf{E}, \tilde{\mathbf{E}}, \hat{\mathbf{E}}\}$,
any strictly proper scoring rule $\ell$ in Eq.~\eqref{eq:loss-kinds} has the conditional
distribution $p^*(\cdot \mid \mathbf{V}, q) = P(Y=\cdot \mid \mathbf{V}, q)$ as its
unique population-risk minimizer. Moreover, because $\tilde{\mathbf{E}}$ is a
deterministic coarsening of $\mathbf{E}$, the entropies of these conditional
distributions satisfy}
\begin{equation}
  H(Y \mid \tilde{\mathbf{E}}, q) - H(Y \mid \mathbf{E}, q)
  = I(Y; \mathbf{E}_T \mid \mathbf{E}_R, q) \ge 0.
  \label{eq:supp-mi}
\end{equation}

\emph{Consequently:}
\begin{enumerate}[nosep, leftmargin=1.5em]
  \item \emph{Non-redundant target evidence ($I(Y; \mathbf{E}_T \mid \mathbf{E}_R, q) > 0$)
        yields strict entropy increase
        $H(Y \mid \tilde{\mathbf{E}}, q) > H(Y \mid \mathbf{E}, q)$.}
  \item \emph{Non-deterministic residual context ($H(Y \mid \tilde{\mathbf{E}}, q) > 0$)
        implies $\max_{m \in \mathcal{Y}(q)} P(Y=m \mid \tilde{\mathbf{E}}, q) < 1$.}
  \item \emph{For control ablation, if $I(Y; \mathbf{E}_c \mid \mathbf{E}_{\setminus c}, q) = 0$,
        then $P(Y \mid \hat{\mathbf{E}}, q) = P(Y \mid \mathbf{E}, q)$.}
\end{enumerate}

\paragraph{Proof.}
We proceed in three steps.

\smallskip\noindent\textit{Step 1: Proper scoring rules.}
For each fixed view $\mathbf{V}$, cross-entropy, the multiclass Brier score, and the
ranked probability score are strictly proper scoring rules whose conditional risk
$\mathbb{E}_{Y\mid \mathbf{V},q}[\ell(p,Y)]$ is uniquely minimized at
$p^*(\cdot \mid \mathbf{V}, q) = P(Y=\cdot \mid \mathbf{V}, q)$.

\smallskip\noindent\textit{Step 2: Entropy identity.}
Note that $\tilde{\mathbf{E}} = (\mathbf{a}, \mathbf{E}_R)$ is a deterministic function
of $\mathbf{E}$, so $H(Y \mid \mathbf{E}, \tilde{\mathbf{E}}, q) = H(Y \mid \mathbf{E}, q)$
and $H(Y \mid \tilde{\mathbf{E}}, q) \ge H(Y \mid \mathbf{E}, q)$; hence
\[
  H(Y \mid \tilde{\mathbf{E}}, q) - H(Y \mid \mathbf{E}, q)
  = I(Y; \mathbf{E} \mid \tilde{\mathbf{E}}, q).
\]
Since $\tilde{\mathbf{E}}$ retains exactly $\mathbf{E}_R$, the chain rule gives
$I(Y; \mathbf{E} \mid \tilde{\mathbf{E}}, q)
= I(Y; \mathbf{E}_T \mid \mathbf{E}_R, q)$,
yielding Eq.~\eqref{eq:supp-mi}.

\smallskip\noindent\textit{Step 3: Consequences.}
Strict positivity of conditional mutual information yields strict entropy increase, and
strictly positive entropy over a finite label set implies that the conditional
distribution is not one-hot, ensuring $\max_m P(Y=m \mid \tilde{\mathbf{E}}, q) < 1$.
Finally, $I(Y;\mathbf{E}_c \mid \mathbf{E}_{\setminus c}, q)=0$ directly implies
$P(Y \mid \hat{\mathbf{E}}, q) = P(Y \mid \mathbf{E}, q)$.
\hfill $\square$

\paragraph{Remark 1 (Transfer to factual calibration under smoothness assumptions).}
Assume that the shared-parameter readout $p_\theta(c \mid \mathbf{V}, q)$ is
$L$-Lipschitz with respect to the full ablated block $\mathbf{E}_T$ (ledger tokens,
bound 2D evidence, and global slot) under a norm $\|\cdot\|$, so that
\[
  |p_\theta(c \mid \mathbf{E}, q) - p_\theta(c \mid \tilde{\mathbf{E}}, q)|
  \le L \|\mathbf{E}_T - \mathbf{a}\|.
\]
When training on the target-ablated view approximately reaches its population minimizer
$p_\theta(c \mid \tilde{\mathbf{E}}, q) \approx P(Y=c \mid \mathbf{E}_R, q)$, combining
the two relations yields
\begin{equation}
  p_\theta(c \mid \mathbf{E}, q)
  \lesssim P(Y=c \mid \mathbf{E}_R, q) + L\|\mathbf{E}_T - \mathbf{a}\|,
  \label{eq:supp-lipschitz}
\end{equation}
up to finite-sample optimization error.

\smallskip\noindent\textit{Interpretation.}
Under smoothness of the readout, joint supervision on factual and target-ablated views
couples factual confidence to the tempered ablated posterior: when the factual target
representation is close to the missing-organ baseline, factual confidence is discouraged
from greatly exceeding the confidence supported by the non-target context
(Table~\ref{tab:cct_abl_calib}).

\smallskip\noindent\textit{Complementary mechanism.}
When withholding all image-derived features degrades the decision score
($r > \breve{r}$), the evidence-blind hinge in Eq.~\eqref{eq:cct-blind} encourages
$\breve{c} \le \operatorname{sg}[c] - m_{\mathrm{blind}}$. Because
$\operatorname{sg}[\cdot]$ stops the gradient through $c$, this term directly lowers
blind confidence relative to the current factual confidence and, through shared
parameters, indirectly discourages evidence-independent, query-only confidence on
factual inputs.

\subsection{2D-3D Feature Fusion in Vision Encoder}
\label{sec:supp-fusion}
    The 2D and 3D channels share organ identity through a deterministic 3D spatial
    membership rule rather than learned alignment. Let the padded volume plane be
    $w_0\times h_0$ and let slice $i$ carry volume index $s_i$. The 2D
    tower emits one patch grid $(1,h_p,w_p)$ per slice. After spatial merging with factor
    $m$, merged cell $(r,c)$, where $0\le r<h_p/m$ and $0\le c<w_p/m$, has its center back-projected
    to original slice coordinates:
    \begin{equation}
      y_r = \left(r+\tfrac{1}{2}\right) m\, w_0 / h_p,
      \qquad
      x_c = \left(c+\tfrac{1}{2}\right) m\, h_0 / w_p .
    \end{equation}
    Write the organ box as $R_k=(z_0,z_1,y_0,y_1,x_0,x_1)$ with inclusive bounds, where an all-$-1$
    row denotes an absent organ. A cell belongs to organ $k$ if and only if
  \begin{equation}
      \begin{aligned}
        M_k(i,r,c) ={}& \mathbb{I}\!\left[z_0\le s_i\le z_1\right]\,
        \mathbb{I}\!\left[y_0\le y_r\le y_1\right] \\
        &\times \mathbb{I}\!\left[x_0\le x_c\le x_1\right].
      \end{aligned}
    \end{equation}

    Thus, slice selection ($s_i\in[z_0,z_1]$) and the in-plane box test act in the same
    coordinate frame, requiring no registration. The hit set
    $\mathbf{S}_k=\{\mathbf{ie}_{i,r,c}: M_k(i,r,c)=1\}$ forms the exact key and value set of
    Eq.~\eqref{eq:evidence}. Specifically, the geometry determines \emph{which} slice tokens represent organ
    $k$, while the residual cross-attention with query $\mathbf{h}_k$ determines their
    weights. When $\mathbf{S}_k$ is empty (e.g., if the sampled slices miss the $z$-range, the box
    lies outside the plane, or the organ is absent), the evidence token falls back to
    $\mathbf{h}_k$ itself, adding zero information rather than fabricating a region.
    Mean-pooling over the same mask yields the parameter-free, organ-level 2D summary consumed by
    the organ-residual variant. Both consumers share this single coordinate convention.
     \begin{figure*}[htp]
        \centering
        \includegraphics[width=0.95\textwidth]{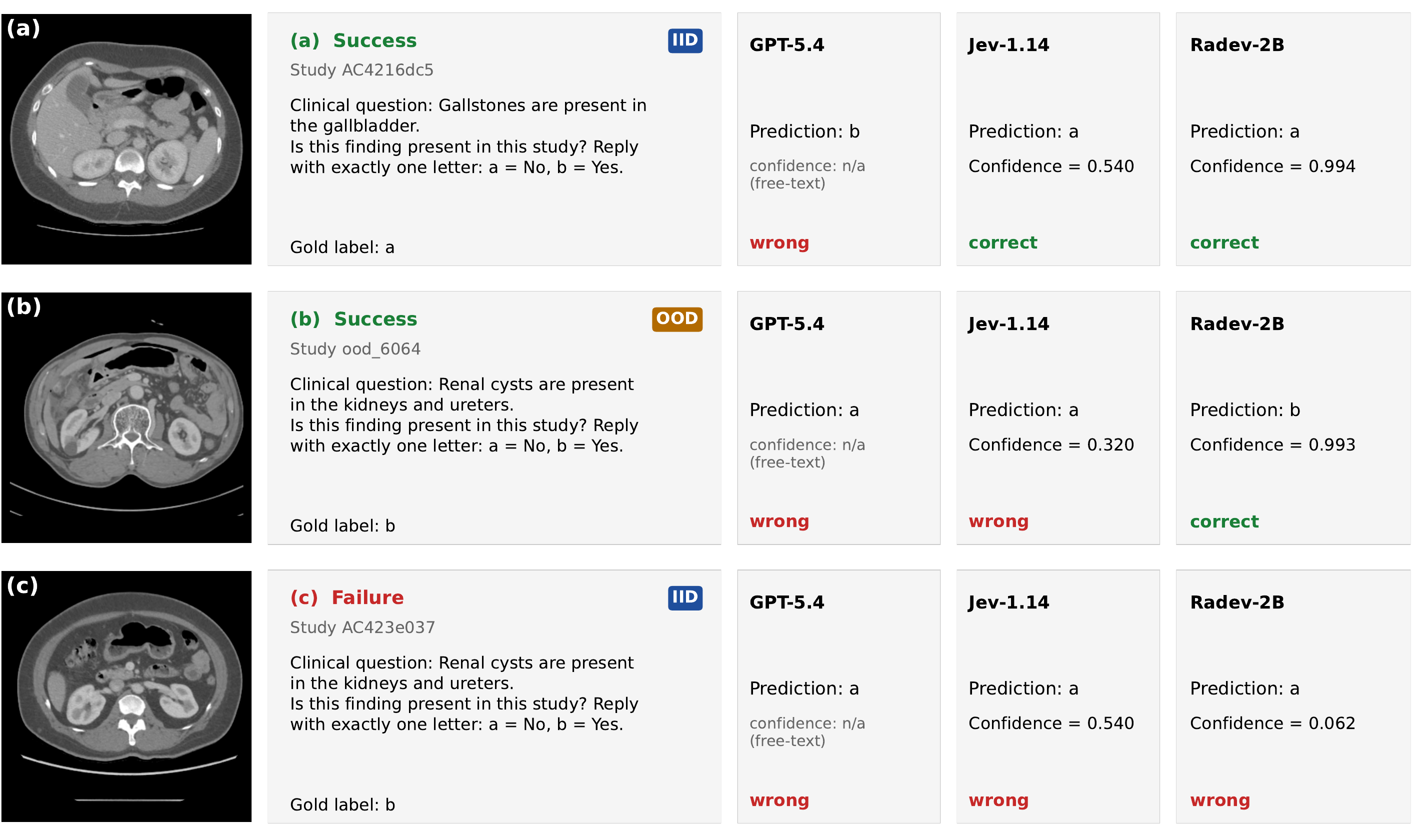}
        \vspace{0.2cm}
        \caption{\textbf{Qualitative verdicts on CT-Verdict-IID and CT-Verdict-OOD.}
        Each row shows one axial slice, the full dataset question with its gold
        label, and the verdicts of GPT-5.4 (free-text), Jev-1.14 (likelihood head),
        and Radev-2B (ours). Rows (a) and (b) are successes of Radev on IID and OOD
        respectively; row (c) is a failure shared by all three models, on which
        Radev reports the lowest confidence.}
        \label{fig:case_study}
    \end{figure*}
 \section{Additional Experimental Results}
 \label{sec:supp-result}
  \subsection{Qualitative Case Study}
\label{sec:case}
Figure~\ref{fig:case_study} illustrates model behavior across both correct and incorrect predictions. On correct cases, Radev-2B consistently outputs decisive confidence (0.994 for an IID study without gallstones in (a); 0.993 for an external AMOS study with renal cysts in (b)), demonstrating that organ-grounded evidence supports reliable verdicts under both in-distribution and distribution-shifted settings. In contrast, baselines either hallucinate positive findings or remain uninformative near chance (e.g., Jev-1.14 at 0.540). Crucially, on a shared failure case where all three models miss a positive renal cyst (c), only Radev-2B's confidence collapses to 0.062, effectively flagging its own error as unreliable. Meanwhile, GPT-5.4 provides no usable score and Jev-1.14 remains uninformative at 0.540, rendering neither error interceptable downstream. This demonstrates that Radev-2B not only errs less overall, but fails loudly: a simple confidence threshold would abstain on the erroneous case (c) while retaining the correct decisions in (a) and (b), converting silent failures into rejectable ones via selective prediction.

\end{document}